# Optical-Guided Neural Collapse for SAR Few-Shot Class Incremental Learning

Fan Zhang, Sijin Zheng, Fei Ma, Qiang Yin, Yongsheng Zhou, Fei Gao, and Xian Sun

Abstract—Few-shot class-incremental learning (FSCIL) in synthetic aperture radar (SAR) imagery presents unique challenges due to severe data scarcity and SAR-specific variability. In particular, strong azimuth sensitivity in SAR induces large intra-class variation and inter-class confusion, and FSCIL's sequential updates further lead to catastrophic forgetting of previously learned classes. Inspired by neural collapse, we propose an optical-guided SAR FSCIL framework, which derives orthogonal feature subspaces from a data-rich optical ATR dataset and uses them as geometric priors to guide SAR feature learning. SAR features are projected onto these orthogonal subspaces via principal angle (PA) constraints, effectively transferring discriminative structure from the optical to the SAR domain. Specifically, our projection loss and the classifier loss optimized with a frozen simplex-ETF geometry jointly induce neural collapse by concentrating features around class means while maintaining large inter-class angles. We evaluate the approach on a benchmark comprising an optical ATR dataset and a SAR ATR dataset with 24 target classes, organized into a base training session and seven incremental sessions. Compared with recent FSCIL methods including NC-FSCIL, PriViLege, WaRP, and FACT, our method achieves the highest final accuracy and a favorable trade-off between final performance and performance degradation. Moreover, neural-collapse metrics show improved intra-class compactness and inter-class separability, indicating that the learned features more closely approximate the ideal simplex-ETF geometry.

***Index Terms*—Few-shot Class-Incremental Learning (FSCIL); Neural Collapse; Orthogonal Subspace Alignment; Synthetic Aperture Radar (SAR)**

## I. INTRODUCTION

SAR serves as a reliable data source for target recognition, as it provides high-resolution imagery under all-day and all-weather conditions [1]. Traditional convolutional neural network (CNN) based SAR target recognition methods perform well on single tasks with sufficient training data. In real-world applications, SAR data often arrives as a stream, where there is a need to continually learn multiple novel SAR targets from separate tasks. When handling such task streams, models often suffer from catastrophic forgetting, which leads to degraded performance on previously learned classes as new classes are introduced [2]. Thus, it remains a challenge to train a model that can incrementally learn from samples of unseen classes without forgetting previously learned ones. This challenge is commonly referred to as class-incremental learning (CIL).

When unseen feature distributions arise in an incremental task stream, the conventional CIL paradigm primarily employs two strategies for adapting to new data: prototype-based and architecture-based approaches. The former involves storing exemplars of previously learned classes and reserving margins for unseen ones [3], while the latter focuses on optimizing network structures via techniques like feature replay [4][5][6], network expanding[7][8], and network distillation[9][10]. However, these strategies are originally designed for scenarios with abundant, balanced data—a condition rarely satisfied in real-world incremental learning, where models must be updated using only small amounts of labeled data. This gives rise to the significant challenge of continuously training models on limited unseen-class data while avoiding forgetting, a task defined as FSCIL.

To tackle data scarcity and imbalance in FSCIL, recent studies have shown that allocating feature space partitions (e.g., orthogonal space [2]) for new and previously learned classes is an effective strategy [2][11][12][14]. Among these algorithms, delicate regularization [12] and dynamic network adjustment strategies [11][14] are employed to preserve the stability of the pre-assigned feature space partitions during the incremental learning process. These methods aim to prevent distributional shifts of previously learned prototypes while maintaining clear separation between the prototypes of old and new classes [15]. Orthogonal feature spaces, where class prototypes are mutually orthogonal, are widely adopted in the literature, as they help reduce inter-class feature confusion and maintain strong class separability under both cosine and Euclidean distance metrics [2]. Therefore, some works introduce orthogonal prototype construction [16] or orthogonality-based loss [17] functions to mitigate the degradation of both class-representative and class-discriminative knowledge during continual learning.

This work was supported in part by the National Natural Science Foundation of China under Grant No.62201027 and No.62331026, Natural Science Foundation of Shandong Province under Grant ZR2024ZD19.

Sijin Zheng, Fei Ma, Qiang Yin, Yongsheng Zhou are with the School of Information Science and Technology, Beijing University of Chemical Technology, Beijing 100029, China (e-mail: 2023200797@buct.edu.cn; mafei@buct.edu.cn; yinq@mail.buct.edu.cn; zhyosh@mail.buct.edu.cn).

Fan Zhang is with the School of Information Science and Technology and the Interdisciplinary Research Center for Artificial Intelligence, Beijing University of Chemical Technology, Beijing 100029, China (e-mail: zhangf@mail.buct.edu.cn).

Fei Gao is with the Beihang University, Beijing 100191, China (e-mail: feigao2000@163.com)

Xian Sun is with the Aerospace Information Research Institute, Chinese Academy of Sciences, Beijing 100190, China (e-mail: sunxian@aircas.ac.cn)

Recent studies have revealed that in the ideal training phase of classification models, when the training error rate reaches zero, a universal geometric structure emerges in the feature space, known as neural collapse [21]. In this regime, last-layer features converge to their respective class centers, which in turn collapse into prototypes forming a simplex equiangular tight frame (ETF) [21]. The ETF geometry provides both strong inter-class separability and intra-class compactness, making it particularly appealing for classification tasks [15][22][23]. In FSCIL, where data distribution is extremely imbalanced, such geometric properties are particularly beneficial: the inherent balance and symmetry of ETFs help mitigate the degradation of class-representative and class-discriminative knowledge [15]. However, strictly converging to a simplex ETF is often overly idealized, as practical feature distributions frequently deviate from this geometry [24]. To bridge this gap, evidence suggests that enforcing orthogonal constraints offers a more structured and practically achievable form of feature collapse [25]. Unlike a simplex ETF, which is mathematically elegant but often unattainable under real-world constraints, an orthogonal frame provides a more structured and achievable geometric target. Importantly, any orthogonal frame can be transformed into a simplex ETF through decentralization and rescaling [25], suggesting that orthogonality can serve as a stepping stone toward the ETF optimum. This insight motivates us to leverage orthogonal constraints as intermediate anchors, ensuring that features not only maintain intra-class compactness but also gradually approach ETF-like geometry in a more stable manner.

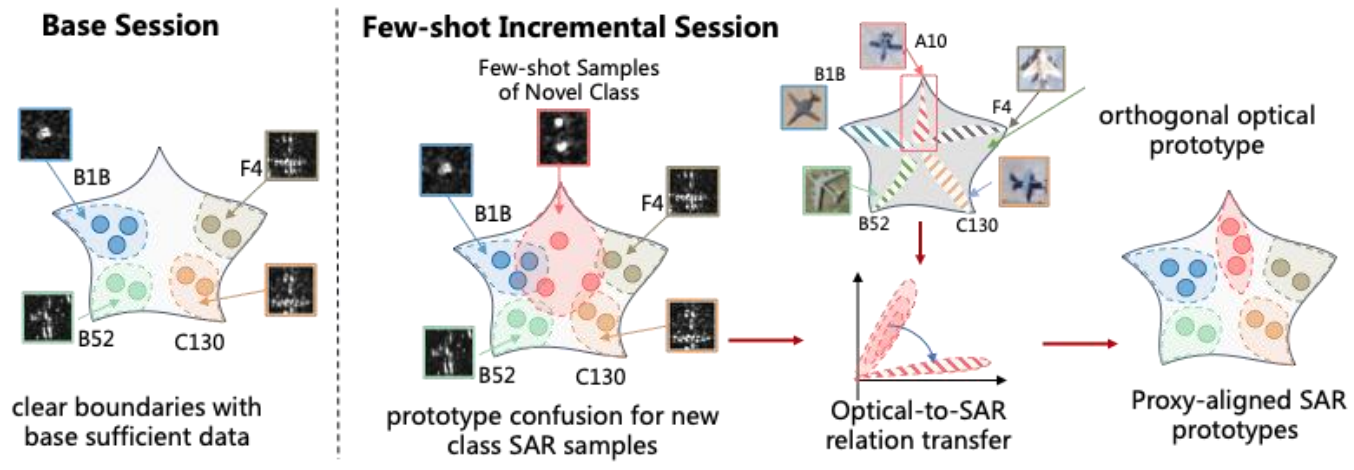

**Fig. 1:** Inspiration for the proposed SAR FSCIL solution

SAR FSCIL presents unique challenges beyond conventional FSCIL. Due to the imaging mechanism of SAR, the same target can exhibit drastically different appearances under different azimuths, making it harder to learn consistent class-representative features. Additionally, visually distinct targets may appear deceptively similar at certain azimuth intervals (e.g., 0°–30° and 150°–180°), which undermines reliable class-discriminative separation. These issues are particularly pronounced in drone-borne and airborne SAR platforms, where flexible flight trajectories result in broader azimuth ranges and hence greater intra-class variability. Compared to SAR, optical imagery is data-rich and well balanced, allowing the formation of orthogonal frame through training and feature construction. By projecting SAR features onto the orthogonal frame, we impose orthogonal constraints on their distributions, thereby reducing the gap between SAR features and the ideal ETF geometry. Building upon this intuition, we propose a neural collapse–inspired framework that couples optical orthogonal frame initialization with cross-domain alignment to enhance stability and generalization in SAR FSCIL.

As illustrated in Fig. 1, feature learning in the data-rich optical domain produces well-separated subspaces with clear decision boundaries. In contrast, few-shot learning in the SAR domain often results in overlapping class boundaries due to limited data. To address this domain gap, we propose to guide SAR feature learning by projecting SAR features into structured subspaces derived from optical data. Specifically, we first construct orthogonal subspaces from a well-trained optical classification model, where each subspace partition serves as a proxy anchor to reinforce the neural collapse of SAR features. To achieve this, we apply Singular Value Decomposition (SVD) to decouple class-specific clusters and compute the inter-class Principal Angles (PAs), selectively enhancing orthogonal bases to maximize subspace separation. Next, to enable effective few-shot learning in the SAR domain, we perform incremental learning guided by optical data: 1) we initialize the classifier weights using a fixed ETF geometry to provide a strong geometric prior; 2) we introduce PA-based orthogonality constraint to drive SAR feature toward its corresponding partitions in the optical orthogonal subspaces—forming the core of our optical-to-SAR relation transfer strategy; 3) a regularized MSE loss to further enhance the convergence of SAR features to the subspaces. These components work together to progressively align SAR features with the neural collapse optimum, ensuring stable learning throughout the incremental process. In summary, our contribution is threefold:

1) **Orthogonal subspace construction**. We propose an optical feature construction method that learns structured, orthogonal subspaces from a well-trained optical classification model, serving as a geometric prior for SAR feature learning.
2) **Optical-to-SAR relation transfer**. We propose an optical-SAR class relation transformation module to gradually transfer the optical remote sensing knowledge into few-shot SAR prototypes via proxy alignment strategy, which minimizes the PA between SAR features and optical subspaces and explicitly steers SAR features toward subspace centers.
3) **Experimental validation**. We conduct extensive experiments on SAR FSCIL benchmarks, demonstrating that our orthogonality constraints help stabilize the simplex ETF structure of SAR features, promoting neural collapse throughout incremental learning.

## II. Related Work

### A. Few-Shot Class-Incremental Learning

This problem was first formalized by Tao et al. [18] in the TOPIC framework, which extends class-incremental learning (CIL) to the few-shot regime. Early solutions like TOPIC leveraged neural gas networks to preserve feature manifold topology, while subsequent methods employed deep-learning techniques, e.g., distillation, to retain past knowledge during learning. Cheraghian et al. [19] transferred critical features via semantic-aware knowledge distillation, while Zhang et al. [20] incorporated calibration modules to balance old/new class scores. Meanwhile, prototype-based methods emerged as an alternative approach, as Zhu et al. [31] enhanced class prototypes through synthesized examples. Also, many subspace-based techniques have been proved effective: Akyürek et al. [12] introduced subspace regularizes to guide new-class weight vectors near existing classes' subspace, and Kim et al. [14] rotates the model's weight space to accommodate new classes without disrupting old decision boundaries, Zhou et al. [30] reserves embedding space for future classes by compressing known-class features and inserting virtual prototype placeholders.

### *B. Neural Collapse and Simplex ETF Geometry*

Papyan et al. [21] first documented neural collapse and its simplex ETF structure at zero training error. Tirer and Bruna [25] extended this finding by proving that with ReLU activations and bias, global minimizers still align features and classifier weights to an ETF structure, making the theory more relevant to practical deep networks. Zhu et al. [23] further proposed fixing classifier weights to a simplex ETF to enhance feature representation. In FSCIL, Yang et al. [15] demonstrated that constraining features to a fixed ETF classifier leads to superior performance compared to orthogonal initialization.

Following Tom et al. [24], methods that initialize prototypes and treat features as free optimization variables are termed unconstrained features models (UFM). Tirer and Bruna [25] analyzed nonlinear deep biased-UFMs, showing that features and weights asymptotically align with a simplex ETF at global minima, and proposed the NC2 metric to quantify this convergence. While these results provide a strong theoretical basis for ETF-based initialization, Tom et al. [24] argued that strict ETF convergence is overly idealized, as practical networks rarely achieve perfect collapse without additional regularization. Since any orthogonal frame can be mean-centered into a simplex ETF [25], Evan et al. [26] introduced orthogonality constraints on classifier weights, demonstrating more effective and faster convergence toward neural collapse than fixed ETF alignment alone.

In contrast to these efforts, our method first constructs optical features to form an orthogonal subspace. By projecting SAR features onto this subspace, we impose orthogonal constraints on their distributions, which in turn drive SAR features closer to ETF geometry and enable more stable neural collapse in class-imbalanced FSCIL scenarios.

## III. PRELIMINARIES

### *A. How Can Neural Collapse Facilitate SAR FSCIL?*

1) Description of neural collapse

In a standard classification task, we denote the input image as $x$ with label $y \in \{1,2,...,C\}$. After passing through the backbone network, the extracted last-layer feature representation of the sample is given by

$$\mu = f(x;\theta_{backbone}) \tag{1}$$

where $\mu \in \mathbb{R}^d$ denotes the d-dimension feature vector of the sample at the final layer. For class $c \in \{1,2,...,C\}$, the class-specific mean feature is defined as $\bar{\boldsymbol{\mu}}_c$, obtained by averaging all features belonging to class $c$. Let $\boldsymbol{\mu}_G$ denote the global mean feature across all classes, then we define the decentralized class center for class $c$ as $\boldsymbol{\mu}_c = (\bar{\boldsymbol{\mu}}_c - \boldsymbol{\mu}_G)^{\mathrm{T}} / \| \bar{\boldsymbol{\mu}}_c - \boldsymbol{\mu}_G \|$. By stacking the decentralized class centers from all $C$ classes, we form a matrix:

$$\mathbf{E} = [\boldsymbol{\mu}_1, \boldsymbol{\mu}_2, ..., \boldsymbol{\mu}_C] \in \mathbb{R}^{d\times C} \tag{2}$$

Neural collapse describes a phenomenon occurring in the final stage of training, once the training error rate reaches zero on balanced data [21]. It reveals a geometric structure correlating to $\boldsymbol{\mu}_c$, which can be characterized by a simplex ETF. An ETF refers to a matrix composed of $C$ vectors in $\mathbb{R}^d$, $d \geq C-1$, i.e.

$$\mathbf{E}_{ETF} = \sqrt{\frac{C}{C-1}}\mathbf{U}\left(\mathbf{I}_C - \frac{1}{C}\mathbf{I}_C\mathbf{I}_C^T\right) \tag{3}$$

where $\mathbf{U} \in \mathbb{R}^{d\times C}$ is a partial orthogonal matrix satisfies $\mathbf{U}^T\mathbf{U} = \mathbf{I}_C$. As defined in equation (2), $\mathbf{E}$ contains $C$ columns of vectors, each representing a decentralized class center vector $\boldsymbol{\mu}_c$ having the same $\ell_2$ norm, and we have:

$$\boldsymbol{\mu}_{c_1}^T\boldsymbol{\mu}_{c_2} = -\frac{1}{C-1}, \forall c_1, c_2 \in [1,C],\ c_1 \neq c_2 \tag{4}$$

where $c_1$ and $c_2$ are any pair of vectors in $\mathbf{E}$.

Let $\boldsymbol{\mu}_{c,l} (c = 1,...,C)$ be the $l$-th feature from class $c$, and $w_c$ denotes the classifier weight of class c, respectively. Under Neural collapse, Papyan et al. [21] observed that there are 4 key phenomena ($\mathcal{NC}1 - \mathcal{NC}4$) describe the behavior of the last-layer feature and classifier during the final phase of training, i.e.

$\mathcal{NC}1$: The last-layer feature variance within the same category approaches zero, i.e., the covariance converges to zero $\Sigma_c \to 0$, where $\Sigma_c = \mathrm{Avg}_l \left\{ (\boldsymbol{\mu}_{c,l} - \boldsymbol{\mu}_c)(\boldsymbol{\mu}_{c,l} - \boldsymbol{\mu}_c)^{\mathrm{T}} \right\}$.

$\mathcal{NC}2$: Each decentralized feature center, centered by the global mean, asymptotically align with a vertex of an ETF, i.e., $\mathbf{E} \to \mathbf{E}_{ETF}$.

$\mathcal{NC}3$: The decentralized means of the last-layer feature gradually converge with the corresponding last-layer classifiers weights, i.e., $\boldsymbol{\mu}_c \to w_c / \|w_c\|$.

$\mathcal{NC}4$: The network's decision-making simplifies to a nearest-class center classifier, where predictions are determined by selecting the mean of last-layer feature having the smallest Euclidean distance to the test example, i.e., $\arg\max_c \langle \boldsymbol{\mu}, w_c \rangle = \arg\min_c \| \boldsymbol{\mu} - w_c \|$, where $\langle\cdot\rangle$ is the inner product operator, and $\boldsymbol{\mu}$ is the input sample feature representation.

Notably, these feature clusters can be regarded as exhibiting an optimized inter-class equiangular separation. When the number of classes of last-layer features becomes very large (i.e., $C \to \infty$), the optimized geometric structure of inter-class equiangular separation reduces to an orthogonal frame, i.e.,

$$\mu_{c_1}^T \mu_{c_2} = 0, \forall c_1, c_2 \in [1, C],\ c_1 \neq c_2 \quad (5)$$

2) Orthogonal Collapse

Studies such as [26] and [24] suggest that, in practical scenarios, particularly under data scarcity and imbalance, the neural collapse is overly idealized. In these settings, the backbone network lacks the flexibility to arbitrarily adjust feature representations to the geometry optimum, whose distributions often deviate significantly from the ETF geometry. As a result, even fixed ETF classifiers [15][22][23] struggle to align these feature distributions with the neural collapse optimum.

In fact, orthogonality is easier to achieve in practice, since any orthogonal frame can be trivially transformed into a simplex ETF by decentralization and scaling [25]. This process can be viewed as orthogonal collapse. By projecting features onto an orthogonal frame, we enforce orthogonality constraints that effectively approximate the ETF geometry and promote stable convergence, i.e., $f_{orth \to ETF}(\cdot)$.

Assuming we have $C$ feature vectors lying in an orthogonal frame with equal $\ell_2$ norm, i.e., $\{\mu_c\}_{c=1}^{C} \in \mathbb{R}^d$. For any pair of vectors, their pairwise inner products equal zero, i.e.,

$$\mu_{c_1}^T \mu_{c_2} = 0, \forall c_1, c_2 \in [1, C],\ c_1 \neq c_2 \quad (6)$$

We define their global mean as $\bar{\mu} = \frac{1}{C}\sum_{c=1}^{C} \mu_c$. The simplex ETF can be constructed by subtracting each sample by the global (decentralizing) and multiply the same scaling factor, i.e.,

$$\hat{\mu}_c = f_{orth \to ETF}(\mu_c) = \sqrt{\frac{C}{C-1}} \times \frac{(\mu_c - \bar{\mu})}{\| \mu_c \|} \quad (7)$$

which leads to a simplex ETF:

$$\hat{\mu}_i^{\mathrm{T}} \hat{\mu}_j = \begin{cases} 1, & i = j, \\ -\frac{1}{C-1}, & i \neq j, \end{cases} \quad \text{and} \quad \sum_{c=1}^{c} \hat{\mu}_i = 0, \quad (8)$$

Evidences show that orthogonal collapse is more structured and achievable compared to a simplex ETF collapse [25]. As is illustrated in Fig. 2, projecting SAR features into orthogonal subspaces imposes an implicit orthogonality constraint, thereby enhancing their neural collapse. These orthogonal subspaces can effectively approximate the simplex ETF prototypes.

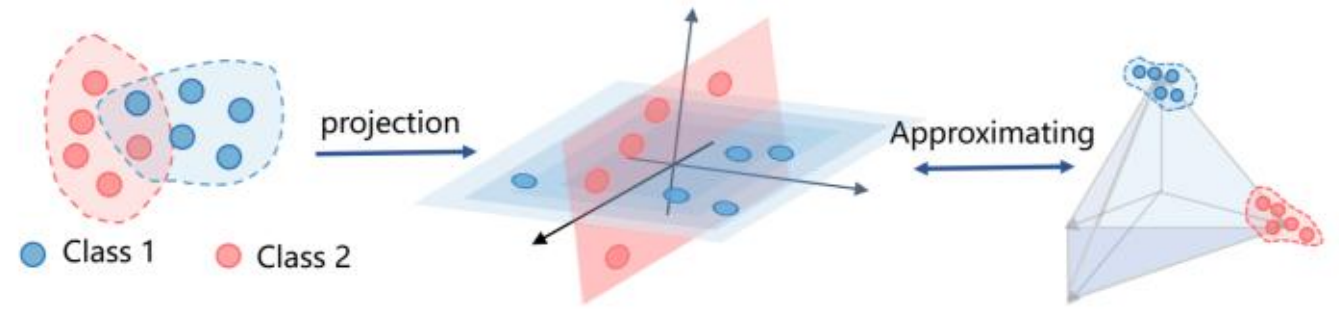

**Fig. 2:** The process of orthogonal collapse. (a) Initial SAR features are projected into (b) orthogonal subspaces, which effectively approximate the (c) simplex ETF prototypes.

In SAR FSCIL, the limited number of SAR samples makes it difficult to construct reliable orthogonal subspaces. In contrast, sufficient optical data enable the formation of class-specific feature clusters that already approximate a simplex ETF. By applying an orthogonal feature construction strategy, these optical features can be transformed into a structured orthogonal frame, which then serves as a proxy anchor to constrain SAR features and guide them toward ETF-like geometry.

## IV. Method

This section proposes a neural collapse-inspired framework for cross-domain FSCIL. As illustrated in Fig. 4, we propose a unified UFM-style architecture across stages: a ViT-based backbone followed by a fixed simplex ETF classifier whose prototype vectors are kept frozen to serve as stable geometric anchors. Subsection IV A introduces the establishment and the construction of orthogonal optical subspaces. In Subsection IV B, we present the detail of our orthogonal subspace alignment strategy with theoretical supports. In subsection IV C, we introduce the network design and training paradigm of the proposed FSCIL framework.

### *A. Orthogonal Optical Subspace Construction*

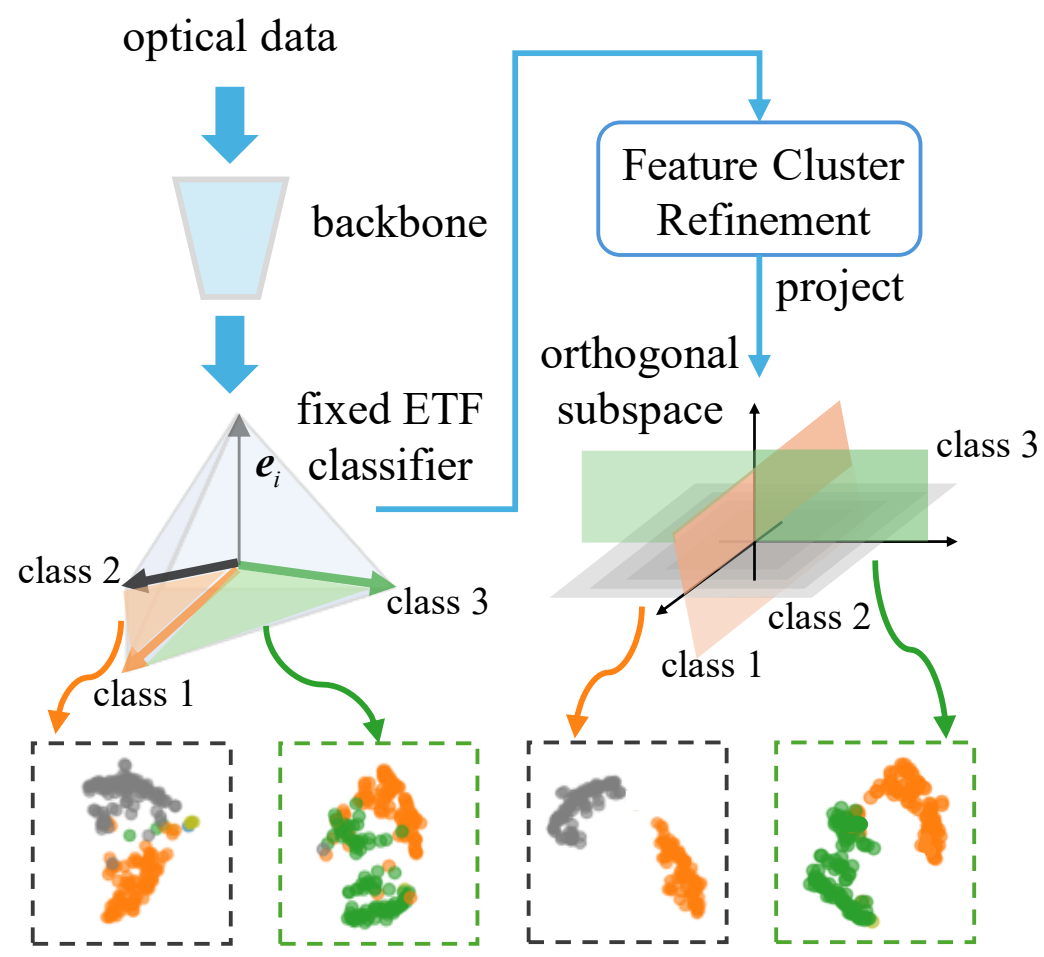

**Fig. 3:** Illustration of orthogonal optical subspace construction

In our cross-domain FSCIL framework, training is divided into two stages: an auxiliary task in the optical domain (source domain) and the main FSCIL task in the SAR domain (target domain).

The key idea is to transfer class-level geometry rather than image content: optical data are used to estimate stable class subspaces, and these subspaces then regularize SAR features when only a few SAR samples are available.

1) Auxiliary optical task (source domain)

Let $\mathcal{D}^{\mathrm{OPT}} = \{x_{c,i}^{opt}\}_{i=1,\ldots,n_c}^{c=1,\ldots,C_{opt}}$ denote the optical ATR dataset, where $x_{c,i}^{opt} \in \mathcal{X}^{\mathrm{OPT}}$ are $i$-th training images of $c$-th class. The model is trained on $\mathcal{D}^{\mathrm{OPT}}$ as a $C_{opt}$-class classification task to establish a feature representation foundation.

2) Main SAR FSCIL task (target domain)

An SAR FSCIL model sequentially receives $N+1$ training session, i.e.,. $\mathcal{D}_0^{\mathrm{SAR}}$ represents the base session and $\mathcal{D}_n^{\mathrm{SAR}} (n = 1, \ldots, N)$ denotes the $n$-th incremental session. Similarly, $\mathcal{D}_n^{\mathrm{SAR}}$ contains $|\mathcal{D}_n^{\mathrm{SAR}}|$ SAR images $x_{c,i}^{SAR}$, i.e., $\mathcal{D}^{\mathrm{SAR}} = \{x_{c,i}^{SAR}\}_{i=1,\ldots,n_c}^{c=1,\ldots,C^{SAR}}$. The base session $\mathcal{D}_0^{\mathrm{SAR}}$ contains a sufficient number of labeled examples, while each incremental session $\mathcal{D}_n^{\mathrm{SAR}}(n=1,...,N)$ introduces a novel class. Importantly, the label sets across all sessions are disjoint. During incremental learning, data from session $n$ is only accessible when learning that session.

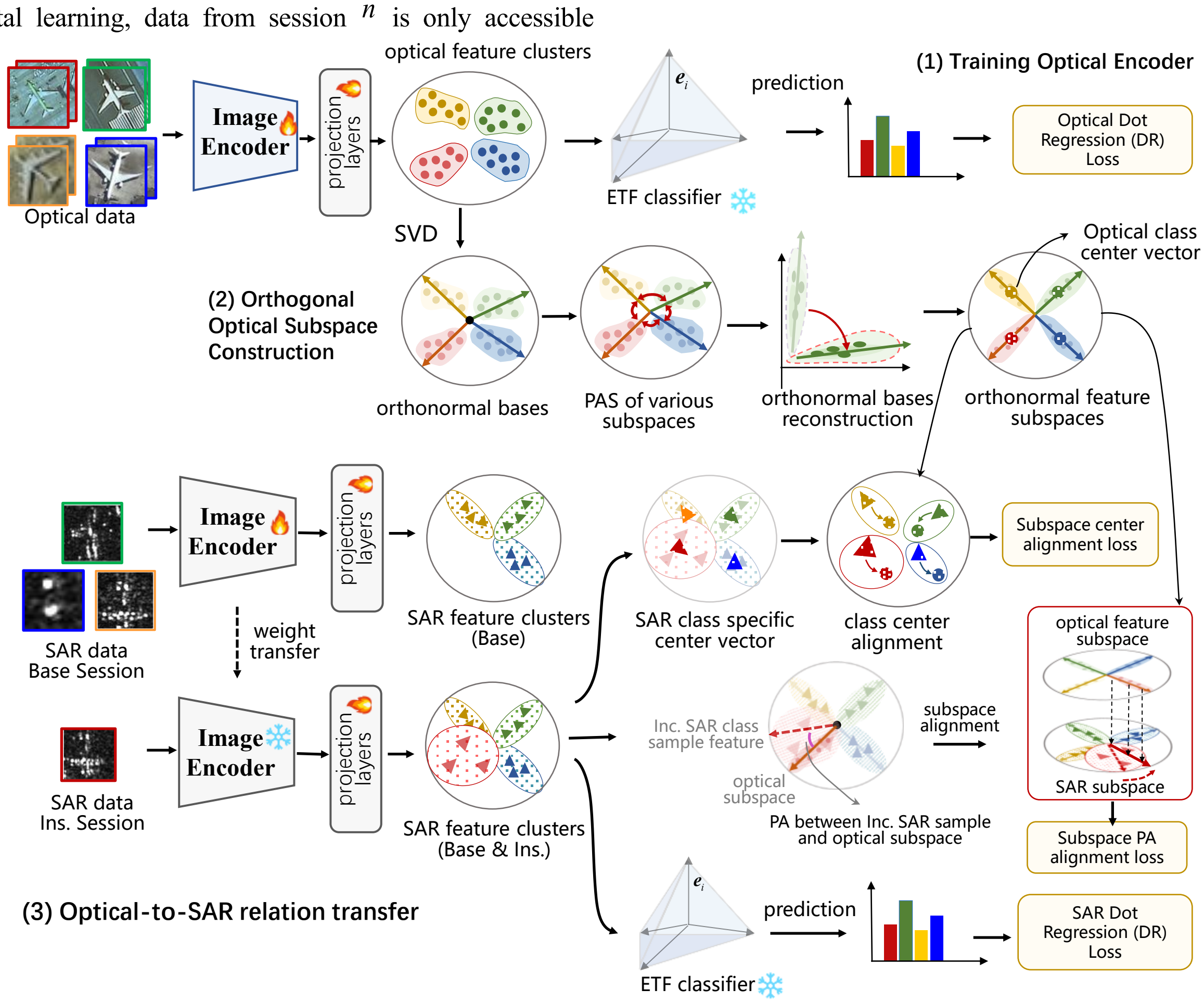

**Fig. 4**: Overall FSCIL framework. (1) Optical samples are encoded and trained with a fixed simplex ETF classifier using optical DR loss to form well-structured class feature clusters. (2) Each optical feature cluster is decomposed to extract an orthonormal basis, while PAS quantifies inter-class subspace orthogonality to guide feature reconstruction, yielding orthogonal subspaces as transferable geometric priors. (3) SAR base classes are learned with DR loss under the same fixed ETF classifier, while each incremental SAR class is guided by aligning features to an unsupervisedly assigned optical subspace using DR loss, subspace PA alignment loss and subspace center alignment loss, with the pretrained image encoder frozen during SAR-domain adaptation.

At the final stage of optical training, the optical feature clusters closely approximate a simplex ETF structure. As discussed in Section III A and illustrated in Fig. 3, we further refine these clusters into structured orthogonal subspaces. This refinement increases inter-class angular separation, removes redundant directions, and produces compact class-level geometric priors for subsequent SAR learning.

We simplify the network into two main modules: a backbone and a classifier head. Since there are $C_{opt}$ classes in the optical domain, the backbone first extracts feature vector $\{\mathbf{f}_i^c\}_{i=1,n_c}^{c=1,C_{opt}} \in \mathbb{R}^d$ from optical ATR image $x_i^{opt}$, where $d$ denotes the feature dimension, $n_c$ is the number of samples for class $c$, and $\mathbf{f}_i^c$ represents the feature of the $i$-th sample belonging to class $c$. Studies [15][21][24] on neural collapse have theoretically demonstrated that under the ideal condition of neural collapse, at the final phase of training, the feature $\mathbf{f}_i^c$ of each sample will converge to its prototype $\mathbf{w}_c^{opt}$, and all class prototypes can converge to a simplex ETF structure, i.e.

$$\mathbf{w}_{c_1}^{ETF\top}\mathbf{w}_{c_2}^{ETF} \propto -\frac{1}{C_{opt}-1}, \quad \forall c_1, c_2 \in [1, C_{opt}] \text{ and } c_1 \neq c_2 \tag{9}$$

The classifier head computes the cosine similarity between the feature $\mathbf{f}_i^c$ and its class prototype $\mathbf{w}_c^{ETF}\left(c=1,...,C_{opt}\right)$. The class corresponding to the prototype with the highest cosine similarity is assigned as the predicted label of the input sample. When a deep classification network reaches the state of neural collapse, which is characterized by intra-class feature compactness and inter-class prototype orthogonality, the classifier naturally evolves into an ETF classifier $\mathbf{W}_{\mathrm{ETF}}$, whose definition is as follows:

$$\begin{aligned} \mathbf{W}^{ETF} &= [\mathbf{w}_1^{ETF},...,\mathbf{w}_{C_{opt}}^{ETF}] \in \mathbb{R}^{d\times C_{opt}} \\ \mathbf{w}_{c_1}^{ETF\top}\mathbf{w}_{c_2}^{ETF} &= -\frac{1}{C_{opt}-1}, \forall c_1, c_2 \in [1, C_{opt}],\ c_1 \neq c_2 \end{aligned} \tag{10}$$

where $\hat{\mathbf{w}}_c^{ETF}$ is the classifier weight vector (class prototype) for the $c$-th class. In practice, these feature clusters can only form an imperfect approximation of the ideal simplex ETF structure [25]. Our objective is to construct orthogonal subspaces in this "imperfect" optical feature space, providing prior information about class relationships for SAR few-shot incremental learning. All $n_c$ features of class $c$ can be formulated as a feature matrix, i.e., $\mathbf{f}_c^{opt} = \left[\mathbf{f}_1^c,...,\mathbf{f}_{n_c}^c\right] \in \mathbb{R}^{n_c \times d}$, where each column corresponds to the $d$-dimensional feature vector of an individual sample in class $c$.

Building on the feature matrices, our strategy aims to construct discriminative orthogonal subspaces by leveraging Singular Value Decomposition (SVD) and Pairwise Alignment (PA). First, we perform SVD on each class-specific feature matrix to extract its orthogonal structural components. The standard SVD of a matrix is formulated as $\mathbf{X} = \mathbf{U}\boldsymbol{\Sigma}\mathbf{V}^{\mathrm{T}}$, where $\mathbf{U}$ consists of orthonormal columns that form the orthogonal basis of the subspace spanned by $\mathbf{X}$ and $\boldsymbol{\Sigma}$ is a diagonal matrix, with diagonal elements (singular values) quantifying the contribution of each corresponding basis vector in $\mathbf{U}$ to the subspace $\mathbf{X}$. For any pair of distinct class feature matrices (e.g. $\mathbf{f}_p^{opt}$ for class $p$ and $\mathbf{f}_q^{opt}$ for class $q$, their SVD decompositions are given by:

In this step, SVD is used only to extract the dominant basis directions of each optical class. The subsequent PAS weighting selects directions that are better separated from other class subspaces, making the transferred optical prior more discriminative for SAR FSCIL.

$$\begin{aligned} \mathbf{f}_p^{opt} &= \mathbf{U}_p^{opt}\boldsymbol{\Sigma}_p^{opt}\mathbf{V}_p^{opt\mathrm{T}} \\ \mathbf{f}_q^{opt} &= \mathbf{U}_q^{opt}\boldsymbol{\Sigma}_q^{opt}\mathbf{V}_q^{opt\mathrm{T}} \end{aligned} \tag{11}$$

where $\mathbf{U}_p^{opt}$ contains the orthonormal bases forming feature subspace $\mathbb{V}_p^{opt}$, similarly, $\mathbf{U}_q^{opt}$ contains the orthonormal bases forming feature subspace $\mathbb{V}_q^{opt}$. $\boldsymbol{\Sigma}_p^{opt}$ and $\boldsymbol{\Sigma}_q^{opt}$ are sequences of singular values, which reflect the importance of each basis vector in their respective subspaces.

To focus on the most expressive dimensions of the class feature space, we truncate matrices $\mathbf{U}_p^{opt}$ and $\mathbf{U}_q^{opt}$'s basis vectors by retaining only their top 5% singular vectors. This yields reduced basis $\tilde{\mathbf{U}}_p^{opt}$ with its singular values $\tilde{\boldsymbol{\Sigma}}_p^{opt}$ for class $p$, analogously, reduced basis $\tilde{\mathbf{U}}_q^{opt}$ with its singular values $\tilde{\boldsymbol{\Sigma}}_q^{opt}$ for class $q$. Then, we utilize the PA to quantify the angular relationship between basis vectors. Specifically, the PA between each basis vector $\tilde{\mathbf{u}}_{p,i}^{opt} \in \tilde{\mathbf{U}}_p^{opt}$ (the $i$-th column of $\tilde{\mathbf{U}}_p^{opt}$ and each basis vector $\tilde{\mathbf{u}}_{q,j}^{opt} \in \tilde{\mathbf{U}}_q^{opt}$ (the $j$-th column of $\tilde{\mathbf{U}}_q^{opt}$) as:

$$\mathbf{PA}(\tilde{\mathbf{u}}_{p,i}^{opt}, \tilde{\mathbf{u}}_{q,j}^{opt}) = \arccos\left(\frac{\left|\tilde{\mathbf{u}}_{p,i}^{opt\mathrm{T}}\tilde{\mathbf{u}}_{p,i}^{opt}\right|}{\left\|\tilde{\mathbf{u}}_{p,i}^{opt}\right\| \cdot \left\|\tilde{\mathbf{u}}_{p,i}^{opt}\right\|}\right) \propto \tilde{\mathbf{u}}_{p,i}^{opt\mathrm{T}}\tilde{\mathbf{u}}_{q,j}^{opt} \tag{12}$$

where $\tilde{\mathbf{u}}_{p,i}^{opt\mathrm{T}}\tilde{\mathbf{u}}_{p,i}^{opt}$ denotes the dot product of $\tilde{\mathbf{u}}_{p,i}^{opt}$ and $\tilde{\mathbf{u}}_{p,i}^{opt}$, and $\|\bullet\|$ represents the Euclidean norm. Consistent with the orthogonal subspace property, $\mathbf{PA}(\tilde{\mathbf{u}}_{p,i}^{opt}, \tilde{\mathbf{u}}_{q,j}^{opt})$ measures the angle between the two basis vectors. $\mathbf{PA}(\tilde{\mathbf{u}}_{p,i}^{opt}, \tilde{\mathbf{u}}_{q,j}^{opt})$ approaches $\pi/2$ when the vectors are nearly orthogonal. Further, the pairwise PAs between subspace $\mathbb{V}_p^{opt}$ and $\mathbb{V}_q^{opt}$:

$$\mathbf{PA}_{pq} = \arccos\left((\tilde{\mathbf{U}}_p)^{\mathrm{T}}\tilde{\mathbf{U}}_q\right) \tag{13}$$

whose element $\mathbf{PA}_{pq}(i,j) = \arccos\left(\tilde{\mathbf{u}}_{p,i}^{opt\mathrm{T}}\tilde{\mathbf{u}}_{q,j}^{opt}\right)$ represent the angle between the $i$-th basis vector of class p and the $j$-th basis vector of class q. In our analysis, we do not restrict attention to identically indexed bases. Instead, we consider all pairwise interactions between the orthonormal bases of classes $p$ and $q$. Every entry of $\mathbf{PA}_{pq}$ quantifies the angular separation between all pairs of subspace directions across the two classes. The collection of angles between orthonormal bases $\left\{\mathbf{PA}_{pq}(i,j)\right\}$ serves as a detailed indicator of orthogonality between feature subspaces $\mathbb{V}_p$ and $\mathbb{V}_q$.

For feature cluster $\mathbf{f}_p^{opt}$ we integrate the angles between each of its orthonormal bases $\tilde{\mathbf{U}}_p^{opt}$ and all bases of class $q$. Specifically, for the $i$-th basis vector $\tilde{\mathbf{u}}_{p,i}^{opt}$, we aggregate its angles over all $\tilde{\mathbf{U}}_q^{opt}$ as:

$$\begin{aligned} \phi_i^{p,q} &= \frac{1}{b_q}\sum\nolimits_{j=1}^{b_q}\mathbf{PA}(\tilde{\mathbf{u}}_{p,i}^{opt}, \tilde{\mathbf{u}}_{q,j}^{opt}) \\ \phi^{p,q} &= [\phi_1^{p,q},...,\phi_{b_p}^{p,q}]^{\mathrm{T}} \end{aligned} \tag{14}$$

where $b_p$ and $b_q$ denotes the number of retained bases for class $p$ and $q$, respectively. For each basis of $\tilde{\mathbf{U}}_p^{opt}$, the vector $\phi^{p,q}$ provides the overall angular separation with respect to the orthonormal bases spanning the feature subspace from $\mathbb{V}_p$ and $\mathbb{V}_q$. When $\phi^{p,q}$ takes large values (i.e., the associated pairwise angles are close to $\pi/2$), the $i$-th orthonormal basis of $\tilde{\mathbf{U}}_p^{opt}$ contributes more to the optimal subspace separation between $\mathbb{V}_p$ and $\mathbb{V}_q$. In contrast, small $\phi^{p,q}$ values indicate that this basis is strongly aligned with the subspace of class $q$ implying stronger inter-class correlation and thus reduced discriminative capability.

To quantitatively evaluate how much each orthonormal basis of $\mathbf{f}_p^{opt}$ contributes to establishing an orthogonal separation between the subspaces $\mathbb{V}_p$ and $\mathbb{V}_q$, we define a Pairwise

Alignment Score (PAS). For the feature cluster $\mathbf{f}_p^{opt}$ against $\mathbf{f}_q^{opt}$ PAS is given by:

$$\mathbf{PAS}(\mathbf{f}_p^{opt},\mathbf{f}_q^{opt}) = normalize\left(\sigma_1 \cdot normalize(\tilde{\Sigma}_p^{opt}) \cdot \frac{\phi^{p,q}}{\pi} + \sigma_2 \cdot normalize(\tilde{\Sigma}_p^{opt})\right) \tag{15}$$

where the normalization operation is applied to maintain data interpretation consistency, i.e.

$$normalize(\boldsymbol{x}) = \frac{\boldsymbol{x} - \min(\boldsymbol{x})}{\max(\boldsymbol{x}) - \min(\boldsymbol{x})} \tag{16}$$

As in equation (15), for the identified 'valuable' orthonormal bases, we assign higher scores.

In the actual feature cluster construction calculation, we iterate over all optical feature clusters $\{\mathbf{f}_1^{opt},...,\mathbf{f}_{C_{opt}}^{opt}\}$, calculating their pairwise **PAS** i.e.

$$S(\mathbf{f}_c^{opt}) = \frac{1}{C_{opt}-1}\sum_{j=1,j\neq i}^{C_{opt}-1} \mathbf{PAS}(\mathbf{f}_i^{opt},\mathbf{f}_j^{opt}) \tag{17}$$

Finally, we further weight $S(\mathbf{f}_c^{opt})$ with the singular values $\tilde{\Sigma}_c^{opt}$ to reconstruct the feature cluster $\mathbf{f}_c^{opt}$ and optimize the orthogonality of feature subspace $\mathbb{V}_c^{opt}$, i.e.,

$$\hat{\mathbf{f}}_c^{opt} = \tilde{\mathbf{U}}_c^{opt} \cdot diag(S(\mathbf{f}_c^{opt}) \cdot \tilde{\Sigma}_c^{opt\mathrm{T}}) \cdot \left(\mathbf{V}_c^{opt}\right)^{\mathrm{T}} \tag{18}$$

where $\hat{\mathbf{f}}_c^{opt}$ is the reconstructed optical feature cluster.

Algorithm 1 summarizes the implementation of the proposed feature-cluster construction strategy.

Fig. 5 visualizes the construction of class-specific optical subspaces. In Fig. 5(a), the unprocessed feature clusters show noticeable overlap and irregular shapes, indicating limited inter-class discriminability. After orthogonal construction, Fig. 5(b) shows more compact and better separated clusters. This comparison supports the need for orthogonal subspace construction before transferring optical geometry to the SAR FSCIL stage.

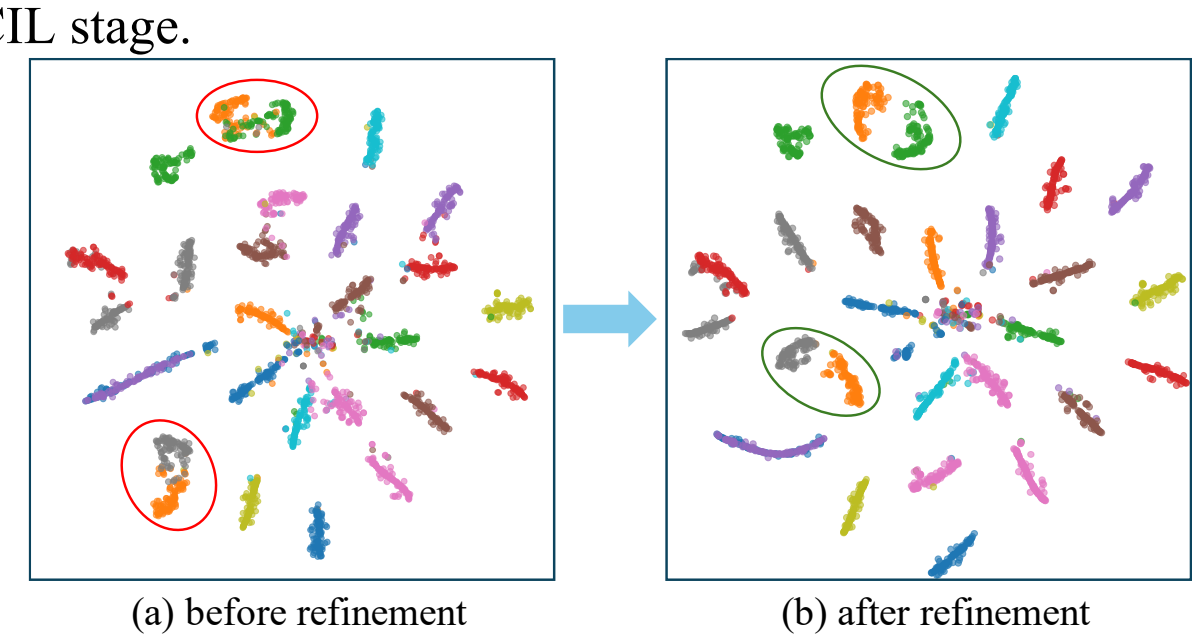

**Fig. 5**: Before-and-after t-SNE visualization of optical feature cluster construction. The reconstructed clusters show clearer inter-class separation and more compact intra-class distributions.

### *B. Optical-to-SAR Relation Transfer*

As discussed in Section III A, we align SAR features not only with their corresponding ETF prototypes but also with the transferred orthogonal optical subspaces. This projection gives SAR features a more stable path toward a simplex ETF-like geometry. However, even when optical and SAR class centers can be roughly aligned, their underlying feature subspaces may still be geometrically mismatched. As illustrated in Fig. 6, we design two complementary alignment strategies: (1) subspace PA alignment, which pulls SAR features toward the corresponding optical subspace, and (2) subspace-center alignment, which stabilizes the features by aligning them with the optical cluster center. Together, these constraints reduce cross-domain subspace mismatch and promote stable convergence toward the neural-collapse optimum. In practice, the three losses play distinct roles. The DR loss keeps features aligned with fixed ETF prototypes, the PA loss pulls incremental SAR features toward the assigned optical subspace, and the center-alignment loss reduces residual fluctuation within that subspace.

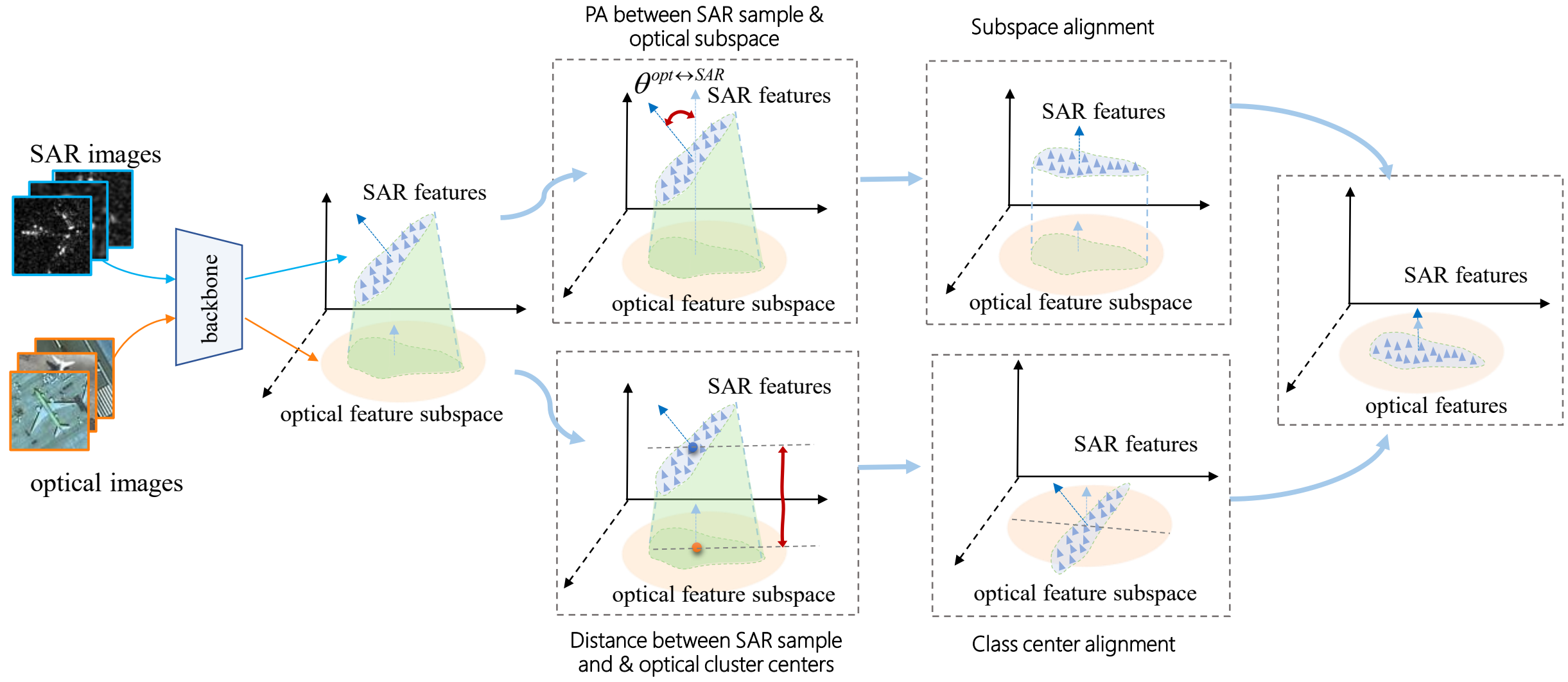

**Fig. 4:** Cross-domain subspace misalignment and our optical-to-SAR relation transfer.

Our SAR FSCIL training follows the UFM formulation, in which the classifier is fixed as a simplex ETF frame while the backbone generates last-layer features to be optimized across both base and incremental sessions. After constructing class-specific orthogonal optical subspaces in Sec. IV A, we leverage these subspaces as geometric priors to guide feature learning in the SAR domain. The process begins with SAR base-class learning. Let $C_{SAR}-1$ denotes the number of SAR base classes. For the $c$ -th class, we denote the $i$ -th class SAR sample and its last-layer feature as $X_{c,i}^{SAR}$ and $\boldsymbol{h}_{c,i}^{SAR}$ , respectively. The backbone extracts the feature $\boldsymbol{h}_{c,i}^{SAR}$ and classification is performed using the fixed ETF prototypes $\mathbf{w}_c^{ETF}\left(c=1,...,C_{opt}\right)$ transferred from the auxiliary optical task in the source domain, as introduced in section II B. Because the base session provides ample data, with a fixed ETF classifier $\mathbf{w}_c^{ETF}$ , a single loss term is enough to drive the features to converge toward the neural-collapse ETF structure. The empirical risk for SAR base-class learning is defined as the average DR loss over all samples:

$$\min_{\widehat{\boldsymbol{h}}_{c,i}^{SAR}\in\mathbb{R}^{d\times N}}\sum_{X_{c,i}^{SAR}\in\mathcal{D}_{base}^{SAR}}^{\left|\mathcal{D}_{base}^{SAR}\right|}\mathcal{L}_1\left(\boldsymbol{h}_{c,i}^{SAR},\mathbf{w}_c^{ETF}\right) \tag{19}$$

where $\mathcal{D}_{base}^{SAR}$ denotes the set of SAR samples available in the base session. We define $N$ as the total amount of training samples, and assume that the global minimizer of equation (19) is $\widehat{\boldsymbol{h}}_{c,i}^{SAR}$ . This optimization target can be realized via DR loss [15]:

$$\mathcal{L}_{DR}=\frac{1}{2}\left(\frac{(\mathbf{w}_c^{ETF})^T}{\left\|(\mathbf{w}_c^{ETF})^T\right\|}\boldsymbol{h}_{c,i}^{SAR}-1\right)^2, \tag{20}$$

Yang et al. theoretically proved that, under both the cross-entropy and dot regression (DR) loss function, the last-layer features converge to a fixed simplex ETF frame. This alignment drives the network toward the global minimizer $\hat{\mu}^{SAR}$, thereby displaying the neural collapse optimum with a single loss setting.

After the base session, our FSCIL protocol proceeds through incremental sessions, each adding a single novel SAR class. Because only the newly added data are available, the feature space becomes unstable and tends to collapse toward previously learned directions. To mitigate this, we transfer structural knowledge from the optical domain by linking each SAR class to the orthogonal optical subspace built in Sec. IV A, and computing the PA between each incremental SAR feature and this subspace. The PA serves as a cross-domain signal, and the refined optical clusters act as proxy anchors that convey their ETF-like geometry to few-shot SAR features, guiding the SAR feature toward a stable orthogonal-collapse configuration.

Because the optical and SAR label spaces are only partially overlapping, we do not assume a one-to-one semantic correspondence between SAR classes and optical classes. Instead, we use a class-agnostic random proxy assignment: before SAR-domain training, each SAR class is randomly assigned to one optical subspace with a fixed random seed, and this assignment remains unchanged across base and incremental sessions. If a SAR class has no corresponding optical category, it is handled in the same way by using the assigned optical subspace as a geometric proxy rather than as a semantic label match. This makes the optical-to-SAR transfer strategy reproducible while keeping the method independent of manual class matching.

At each incremental step, the newly added class is indexed as $C_{SAR}$ . Building on Sec. IV A, where the auxiliary optical task yields a feature cluster $\hat{\mathbf{f}}_{C_{SAR}}^{opt}$ whose centroid is close to the corresponding ETF classifier prototype $\mathbf{w}_{C_{SAR}}^{ETF}$ and spans an optical feature subspace $\mathbb{V}_{C_{SAR}}^{opt}$ , we extrapolate the principal angle between each unseen SAR feature $\boldsymbol{h}_{C_{SAR},i}^{SAR}$ and $\mathbb{V}_{C_{SAR}}^{opt}$ . To jointly enforce alignment with the ETF prototype and the optical subspace, the empirical risk in the incremental stage can be formed as:

$$\min_{\hat{\boldsymbol{h}}_{C_{SAR},i}^{SAR}\in\mathbb{R}^{d\times N}}\sum_{X_{C_{SAR},i}^{SAR}\in\mathcal{D}_{inc}^{SAR}}^{\left|\mathcal{D}_{inc}^{SAR}\right|}\mathcal{L}_{DR}\left(\boldsymbol{h}_{C_{SAR},i}^{SAR},\mathbf{w}_{C_{SAR}}^{ETF}\right)+\mathcal{L}_{orth}\left(\boldsymbol{h}_{C_{SAR},i}^{SAR},\mathbb{V}_{C_{SAR}}^{opt}\right) \tag{21}$$

where $\mathcal{D}_{inc}^{SAR}$ denotes the set of SAR samples at each incremental step, $X_{C_{SAR},i}^{SAR}$ is the i-th new samples, and $\widehat{\boldsymbol{h}}_{C_{SAR},i}^{SAR}$ is its global minimizer feature.

As is mentioned in (20), we use DR loss as $\mathcal{L}_{DR}$ that pulls $\boldsymbol{h}_{C_{SAR},i}^{SAR}$ toward their ETF directions. To align SAR feature $\boldsymbol{h}_{C_{SAR},i}^{SAR}$ with the optical feature subspace $\mathbb{V}_{C_{SAR}}^{opt}$ , we use a PA-based orthogonal loss as $\mathcal{L}_{orth}$ . where each SAR feature $\boldsymbol{h}_{C_{SAR},i}^{SAR}$ is expressed in the orthogonal basis of the optical subspace $\mathbb{V}_{C_{SAR}}^{opt}$ . In this way, we first calculate the self-covariance of $\hat{\mathbf{f}}_{C_{SAR}}^{opt}$ , i.e. $\mathbb{E}_{\hat{\mathbf{f}}_{C_{SAR}}^{opt}\in\mathbb{V}_c}\left(\hat{\mathbf{f}}_{C_{SAR}}^{opt}\cdot(\hat{\mathbf{f}}_{C_{SAR}}^{opt})^T\right)$. The SVD components forming $\hat{\mathbf{f}}_{C_{SAR}}^{opt}$ can be expressed as:

$$\mathbb{E}_{\hat{\mathbf{f}}_{C_{SAR}}^{opt}\in\mathbb{V}_{C_{SAR}}^{opt}}\left(\hat{\mathbf{f}}_{C_{SAR}}^{opt}\cdot(\hat{\mathbf{f}}_{C_{SAR}}^{opt})^T\right)=\mathbf{U}_{C_{SAR}}^{opt}\boldsymbol{\Sigma}_{C_{SAR}}^{opt}\left(\mathbf{V}_{C_{SAR}}^{opt}\right)^{\mathrm{T}} \tag{22}$$

Here, $\mathbf{U}_{C_{SAR}}^{opt}$ contains orthonormal bases and $\boldsymbol{\Sigma}_{C_{SAR}}^{opt}$ represents a set of singular values. We have $\mathbf{R}_{C_{SAR}}^{opt}$ contains the first $D$ columns of $\mathbf{U}_{C_{SAR}}^{opt}$ forming an orthonormal basis for the optical subspace of class $C_{SAR}$ . Consequently, $\left(\mathbf{R}_{C_{SAR}}^{opt}\right)^{\mathrm{T}}\boldsymbol{h}_{C_{SAR},i}^{SAR}$ gives the coordinates of $\boldsymbol{h}_{C_{SAR},i}^{SAR}$ in this basis, and multiplying these coordinates back by $\mathbf{R}_{C_{SAR}}^{opt}$ reconstructs the closest point to $\boldsymbol{h}_{C_{SAR},i}^{SAR}$ inside the subspace. So, the projection vector $\boldsymbol{h}_{C_{SAR},i}^{proj}$ of feature $\boldsymbol{h}_{C_{SAR},i}^{SAR}$ on the subspace $\mathbb{V}_{C_{SAR}}^{opt}$ can be expressed as:

$$\boldsymbol{h}^{proj}_{C_{SAR},i} = \left(\mathbf{R}^{opt}_{C_{SAR}}\right)\left(\mathbf{R}^{opt}_{C_{SAR}}\right)^{\mathrm{T}} \boldsymbol{h}^{SAR}_{C_{SAR},i} \tag{23}$$

The cosine between $\boldsymbol{h}^{proj}_{C_{SAR},i}$ and $\boldsymbol{h}^{SAR}_{C_{SAR},i}$ defines a principal angle that quantifies how well the SAR feature follows the geometry of the optical subspace $\mathbb{V}^{opt}_{C_{SAR}}$, i.e.,

$$\cos(\theta^{\boldsymbol{h}^{SAR}_{C_{SAR},i} \leftrightarrow \mathbb{V}^{opt}_{C_{SAR}}}) = \frac{\left\| \boldsymbol{h}^{proj}_{C_{SAR},i} \right\|}{\left\| \boldsymbol{h}^{SAR}_{C_{SAR},i} \right\|} \tag{24}$$

Let SAR feature $\boldsymbol{h}^{SAR}_{C_{SAR},i}$ be the optimization variable, we then introduce an orthogonal alignment loss that penalizes this angle. The empirical risk of $\mathcal{L}_{orth}$ can be formed as:

$$\mathcal{L}_{orth} \triangleq \min_{\boldsymbol{h}^{SAR}_{C_{SAR},i} \in \mathbb{R}^{d\times N}} \sum_{X^{SAR}_{C_{SAR},i} \in \mathcal{D}^{SAR}_{inc}}^{\left|\mathcal{D}^{SAR}_{inc}\right|} \arccos\left( \frac{\left\| \left(\mathbf{R}^{opt}_{C_{SAR}}\right)^{T} \cdot \boldsymbol{h}^{SAR}_{C_{SAR},i} \right\|}{\left\| \boldsymbol{h}^{SAR}_{C_{SAR},i} \right\|} \right) \tag{25}$$

Based on the geometric construction of the $\mathcal{L}_{orth}$, we now study its global minimizer expressed in Eq. (25). Theorem 1 further demonstrates that this loss admits a global minimizer, leading SAR features to collapse onto their class-specific optical subspaces.

**Theorem 1** *Let $\widehat{\boldsymbol{h}}^{SAR}_{C_{SAR},i} \in \mathbb{R}^{d}$ be the global minimizer of Equation (25). Under loss function $\mathcal{L}_{orth}$ in equation (25). Any $\hat{\boldsymbol{h}}^{SAR}_{C_{SAR},i}$ satisfies:*

$$\cos\angle(\widehat{\boldsymbol{h}}^{SAR}_{C_{SAR},i}, \mathbb{V}^{opt}_{C_{SAR}}) \sim 1,$$
$$\left[\bar{\hat{\boldsymbol{\mu}}}_1, ..., \bar{\hat{\boldsymbol{\mu}}}_{C^{SAR}}\right]^{\mathrm{T}} \left[\bar{\hat{\boldsymbol{\mu}}}_1, ..., \bar{\hat{\boldsymbol{\mu}}}_{C^{SAR}}\right] \sim \mathrm{diag}(\|\bar{\hat{\boldsymbol{\mu}}}_1\|, ..., \|\bar{\hat{\boldsymbol{\mu}}}_{C_{SAR}}\|),$$
$$\forall 1 \le c \le C^{SAR}, 1 \le i \le N \tag{26}$$

*where $\bar{\hat{\boldsymbol{\mu}}}_c$ denote the mean feature of class c.*

*Proof.* Following $\mathcal{L}_{orth}$ in Equation (26), $\mathcal{L}_{orth}$ is always non-negative, and $\mathcal{L}_{orth} = 0$ occur only when $\left\| \left(\mathbf{U}^{opt}_{C_{SAR}}\right)^{\mathrm{T}} \boldsymbol{h}^{SAR}_{C_{SAR},i} \right\| \Big/ \left\| \boldsymbol{h}^{SAR}_{C_{SAR},i} \right\| = 1$, $\forall 1 \le k \le K, 1 \le i \le N$. Note that $\mathbf{U}^{opt}_{C_{SAR}}$ is a matrix whose columns are orthogonal basis vectors, each with a norm of 1, as obtained from the SVD decomposition. Thus, we have $\left\|\mathbf{U}^{opt}_{C_{SAR}}\right\| = 1$. In addition, note that $\left\| \left(\mathbf{U}^{opt}_{C_{SAR}}\right)^{\mathrm{T}} \boldsymbol{h}^{SAR}_{C_{SAR},i} \right\| \in [0,1]$ is the projection of feature $\boldsymbol{h}^{SAR}_{C_{SAR},i}$ onto the subspace $\mathbb{V}^{opt}_{C_{SAR}}$ constructed by the matrix $\mathbf{U}^{opt}_{C_{SAR}}$. we hereby have the following:

$\left\| \left(\mathbf{U}^{opt}_{C_{SAR}}\right)^{\mathrm{T}} \boldsymbol{h}^{SAR}_{C_{SAR},i} \right\| = 0$ holds only when $\boldsymbol{h}^{SAR}_{C_{SAR},i}$ is orthogonal to $\mathbb{V}^{opt}_{C_{SAR}}$, i.e. $\cos\angle(\boldsymbol{h}^{SAR}_{C_{SAR},i}, \mathbb{V}^{opt}_{C_{SAR}}) \sim 0$

$\left\| \left(\mathbf{U}^{opt}_{C_{SAR}}\right)^{\mathrm{T}} \boldsymbol{h}^{SAR}_{C_{SAR},i} \right\| = 1$ holds only when $\boldsymbol{h}^{SAR}_{C_{SAR},i}$ is exactly located within $\mathbb{V}^{opt}_{C_{SAR}}$, i.e. $\cos\angle(\boldsymbol{h}^{SAR}_{C_{SAR},i}, \mathbb{V}^{opt}_{C_{SAR}}) \sim 1$

Therefore, there exists a solution $\hat{\boldsymbol{h}}^{SAR}_{C_{SAR},i}$ as a global minimizer for Equation (25). Any feature vector $\hat{\boldsymbol{h}}^{SAR}_{C_{SAR},i}$ lies within its corresponding orthogonal feature subspace $\mathbb{V}^{opt}_{C_{SAR}}$. Thus, we have:

$$\cos\angle(\hat{\boldsymbol{h}}^{SAR}_{C_{SAR},i}, \mathbb{V}^{opt}_{C_{SAR}}) \sim 1, \ \forall 1 \le c \le C^{SAR}, 1 \le i \le N \tag{27}$$

which concludes with $\mathcal{L}_{orth}$ in equation (26).

As $\mathcal{L}_{orth}$ in Equation (26) optimizing the angle between SAR feature $\boldsymbol{h}^{SAR}_{C_{SAR},i}$ and orthogonal subspace $\mathbb{V}^{opt}_{C_{SAR}}$, we observe that SAR feature may still oscillate, even when they are already exactly located inside the corresponding subspace $\mathbb{V}^{opt}_{C_{SAR}}$. To further stabilize the SAR feature $\boldsymbol{h}^{SAR}_{C_{SAR},i}$ into its corresponding feature subspace $\mathbb{V}^{opt}_{C_{SAR}}$, we formulate another sub-problem:

$$\begin{aligned} &\min_{\hat{\boldsymbol{h}}^{SAR}_{C_{SAR},i} \in \mathbb{R}^{d\times N}} \sum_{X^{SAR}_{C_{SAR},i} \in \mathcal{D}^{SAR}_{inc}}^{\left|\mathcal{D}^{SAR}_{inc}\right|} \mathcal{L}_{center}\left(\boldsymbol{\mu}^{SAR}_{c,l}, \boldsymbol{c}^{opt}_{C_{SAR}}\right) \\ &\text{s.t.} \quad \boldsymbol{c}^{opt}_{i}\boldsymbol{c}^{opt}_{j} = \begin{cases} 0, & i = j, \\ 1, & \text{otherwise} \end{cases} \quad (\forall i, j). \end{aligned} \tag{28}$$

where $\boldsymbol{c}^{opt}_{C_{SAR}} \in \mathbb{R}^{d}$ is presumed to be the center vector of the feature cluster $\hat{\mathbf{f}}^{opt}_{C_{SAR}}$, i.e. $\boldsymbol{c}^{opt}_{C_{SAR}} = \frac{1}{n}\sum_{i=1}^{n}\boldsymbol{\mu}^{opt}_{c,n}$, where $\hat{\mathbf{f}}^{opt}_{C_{SAR}} = [\boldsymbol{\mu}^{opt}_{C_{SAR},1}, ..., \boldsymbol{\mu}^{opt}_{C_{SAR},n}]$. Here, we introduce MSE loss to solve the sub-problem formulated in Equation (28):

$$\mathcal{L}_{center}\left(\boldsymbol{h}^{SAR}_{C_{SAR},i}, \boldsymbol{c}^{opt}_{C_{SAR}}\right) = (\boldsymbol{h}^{SAR}_{C_{SAR},i} - \boldsymbol{c}^{opt}_{C_{SAR}})^2 \tag{29}$$

Since $\mathcal{L}_{center}$ in Equation (29) is convex *w.r.t* feature $\boldsymbol{h}^{SAR}_{C_{SAR},i}$, and the center vector lies within the feature subspace constructed by the feature clusters $\hat{\mathbf{f}}^{opt}_{C_{SAR}}$, we conclude that there exist a global minimizer $\hat{\boldsymbol{h}}^{SAR}_{C_{SAR},i}$ that solves both sub-problems in Equations (25) and (28).

The overall formulation is given by:

$$\mathcal{L} = \lambda_{DR}\mathcal{L}_{DR} + \lambda_{orth}\mathcal{L}_{orth} + \lambda_{center}\mathcal{L}_{center}$$
$$= \frac{1}{2}\lambda_{DR}\left(\frac{(\mathbf{w}_{C_{SAR}}^{ETF})^T}{\left\|(\mathbf{w}_{C_{SAR}}^{ETF})^T\right\|}\boldsymbol{h}_{C_{SAR},i}^{SAR} - 1\right)^2 + \lambda_{orth}\arccos\left(\frac{\left\|\left(\mathbf{R}_{C_{SAR}}^{opt}\right)^T \cdot \boldsymbol{h}_{C_{SAR},i}^{SAR}\right\|}{\left\|\boldsymbol{h}_{C_{SAR},i}^{SAR}\right\|}\right) + \lambda_{center}(\boldsymbol{h}_{C_{SAR},i}^{SAR} - \boldsymbol{c}_{C_{SAR}}^{opt})^2 \tag{30}$$

where $\lambda_{DR}$, $\lambda_{orth}$ and $\lambda_{center}$ denote the weight for $\mathcal{L}_{DR}$, $\mathcal{L}_{orth}$ and $\mathcal{L}_{center}$, respectively.

### *C. Implementation Details*

Motivated by Li et al. [28], who analyze transferability through the lens of neural collapse and suggest that more collapsed pretrained representations tend to transfer better, we build our framework on a large-scale pretrained SAM (Segment Anything Model) encoder and preserve its priors as much as possible. The key design goal is to handle the optical-to-SAR domain shift via parameter-efficient adaptation under the fixed simplex ETF geometry. In the auxiliary optical stage, both the image encoder and the projection layers are trained to learn optical class geometry. When switching to SAR base-class learning and subsequent incremental sessions, the pretrained image encoder is frozen, and only the adapter, projection, and LoRA-related parameters are updated for SAR-domain adaptation.

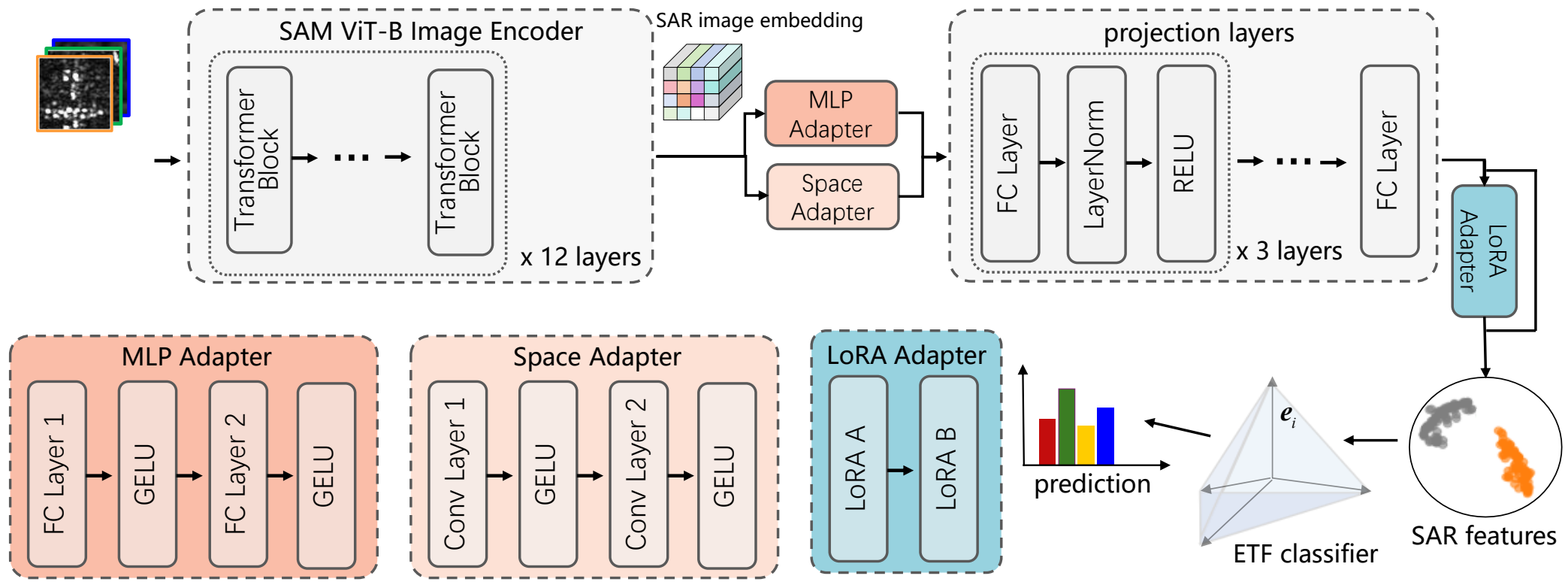

**Fig. 7**: Architecture of the proposed SAM-based encoder and parameter-efficient adaptation module used for SAR FSCIL.

As illustrated in Fig. 7, the backbone is the SAM ViT-B image encoder (12 Transformer blocks) [37]. Two residual adapters are attached near the end of the encoder: the MLP Adapter modifies channel-wise representations, and the Space Adapter adjusts spatial patterns. The resulting features are passed to a projection head consisting of three components, i.e., FC, LayerNorm and ReLU blocks, followed by a final FC layer for representation reshaping. A LoRA Adapter (LoRA A/B) is injected into the final projection layer to enable low-rank updates, while the simplex ETF classifier is kept fixed, providing frozen prototype directions for training.

## V. Experimental Results

This section evaluates the proposed FSCIL method. Subsection V A introduces the datasets and experimental settings, Subsection V B reports the performance comparison, and Subsection V C analyzes the ablation results.

### *A. Dataset and Settings*

We use two datasets: Optical ATR as the source domain and SAR ATR as the target domain. Although their label spaces only partially overlap, the optical ATR dataset contains more categories and richer appearance variations than the SAR ATR dataset. From the perspective of neural collapse, well-trained optical features provide class-specific angular margins and compact class clusters, making them suitable proxy anchors for guiding SAR FSCIL through feature alignment and geometric regularization.

**Optical ATR dataset**: As is shown in the left portion of Fig. 8, several optical ATR datasets are integrated to enhance data diversity. Firstly, the Multi-Type Aircraft Remote Sensing Images (MTARSI) dataset provides 20 categories of aircraft images [34]. These images are collected from 36 airports worldwide and exhibit extensive variations in background scenarios, spatial resolutions, lighting conditions, and times of day. To further enrich the dataset, we collect high-resolution optical remote sensing images of Cessna 208, AT402, and F-4 aircraft from Google Earth. Furthermore, the MAR20 [35] and FAIR1M [36] datasets provide an extra 11 aircraft classes. In total, our optical ATR dataset includes 28 diverse aircraft categories, offering a robust foundation for model initialization and training.

**SAR ATR dataset**: Our SAR ATR dataset integrates four widely recognized SAR datasets, i.e., MSTAR dataset, SAR-AIRcraft-1.0, TerraSAR US Aircraft dataset and UAV-based miniSAR dataset. MSTAR dataset, captured using high-resolution X-band SAR imagery (0.3 m × 0.3 m, HH polarization), includes ten ground mobile targets such as 2S1, ZSU234, and T72. SAR-AIRcraft-1.0 contains GaoFen-3 satellite images (1m resolution) from three airports, covering six aircraft types, including A220, A320, and Boeing 737. The remaining two datasets are described as follows: one is obtained by UAV-mounted miniSAR, collected across multiple polarization modes (HH, HV, VV) and frequency bands (L, C, Ku), containing three aircraft types including YUN12, Cessna 208, and AT402; the other covers five military aircraft classes such as A10, B52, and C130, acquired with X-band SAR imagery at 1 m resolution and HH polarization. As illustrated in the middle (base session) and right (incremental sessions) of Fig. 8, these datasets contribute to a total of 24 SAR target classes, which are divided into base and incremental sessions to support few-shot class-incremental learning on the SAR domain.

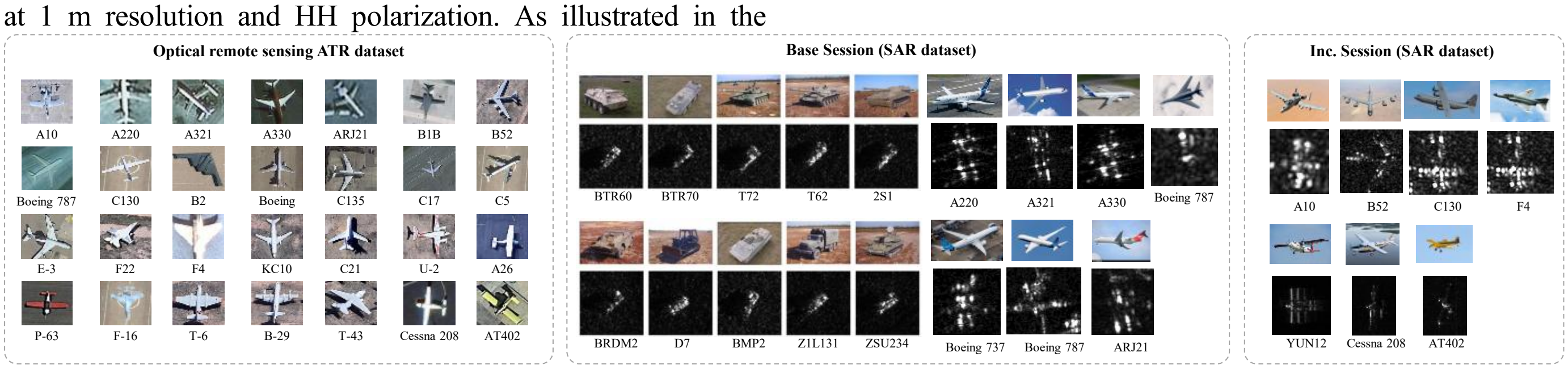

**Fig. 8**: Representative optical images (left), SAR base-session targets (middle), and SAR incremental-session targets (right) used in the cross-domain FSCIL setting.

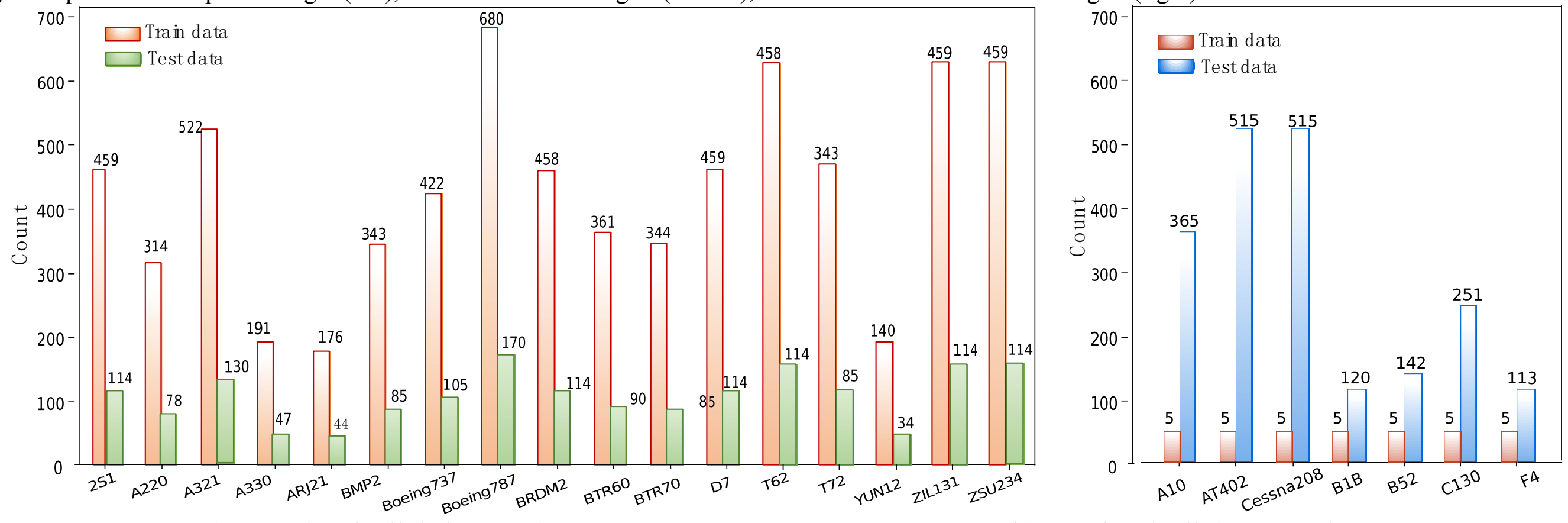

**Fig. 9**: Training and testing sample counts for the SAR dataset, including base-session and incremental-session classes.

Fig. 9 compares the training and testing sample counts for the SAR dataset. Base-session classes are marked in blue, and incremental-session classes are marked in orange, providing a clear overview of the data distribution. In the SAR ATR dataset, novel categories are introduced in a fixed order: A10, AT402, Cessna 208, B1B, B52, C130, and F4, forming seven incremental sessions for stepwise FSCIL evaluation.

### *B. Performance Evaluation*

We first train the model in the optical ATR dataset using the RMSprop optimizer with a learning rate of 0.00005 for 500 epochs. A fixed ETF classifier head is used and the model is optimized with $\mathcal{L}_{DR}$ to encourage feature collapse toward class-specific ETF prototypes. After pre-training, the model achieves a test accuracy of 95.31% on the optical dataset. Owing to the geometric properties of neural collapse, the class-specific feature clusters in the optical feature subspace show strong inter-class separability and intra-class compactness, with their centroids approximately converging toward a simplex ETF structure. For SAR FSCIL, we use our adapter-based transfer strategy. In the base session, all model parameters are fine-tuned for 300 epochs using SGD with a learning rate of 0.08. In incremental sessions, only the projection layers and LoRA adapters are fine-tuned for 50 epochs, with the backbone kept frozen. The training objective in the SAR domain combines the same $\mathcal{L}_{DR}$ with orthogonality regularization terms, i.e., $\mathcal{L}_{orth}$ and $\mathcal{L}_{center}$, which drives the SAR features toward their designated optical subspaces, tthereby promoting more stable and reliable convergence toward their neural collapse minima.

The model is implemented using PyTorch and trained on a platform equipped with an AMD EPYC 7551P CPU and an NVIDIA GeForce RTX A5000 GPU with 24 GB of memory. Specifically, for feature-cluster construction, the hyperparameters $\sigma_1$ and $\sigma_2$ are set to 1 and 2, respectively. During training, the loss weights are configured to 10.0, 8.0, and 2.8, respectively.

We compare the proposed method with several FSCIL algorithms: neural collapse inspired few-shot incremental learning (NC-FSCIL) [15], Pre-trained Vision and Language transformers with prompting functions and knowledge distillation (PriViLege) [13], Weight Space Rotation for Class-Incremental Few-Shot Learning (WaRP) [14], and Forward Compatible Few-Shot Class-Incremental Learning (FACT) [30]. NC-FSCIL fixes classifier weights as equiangular class prototypes covering all base and incremental classes and trains the feature extractor to align its outputs with these static prototypes. PriViLege uses a large pre-trained vision-language transformer with prompt tuning and knowledge distillation for few-shot class increments. WaRP preserves prior knowledge by freezing important parameters and fine-tuning the remaining parameters for new classes. FACT reserves feature space for future classes during incremental learning.

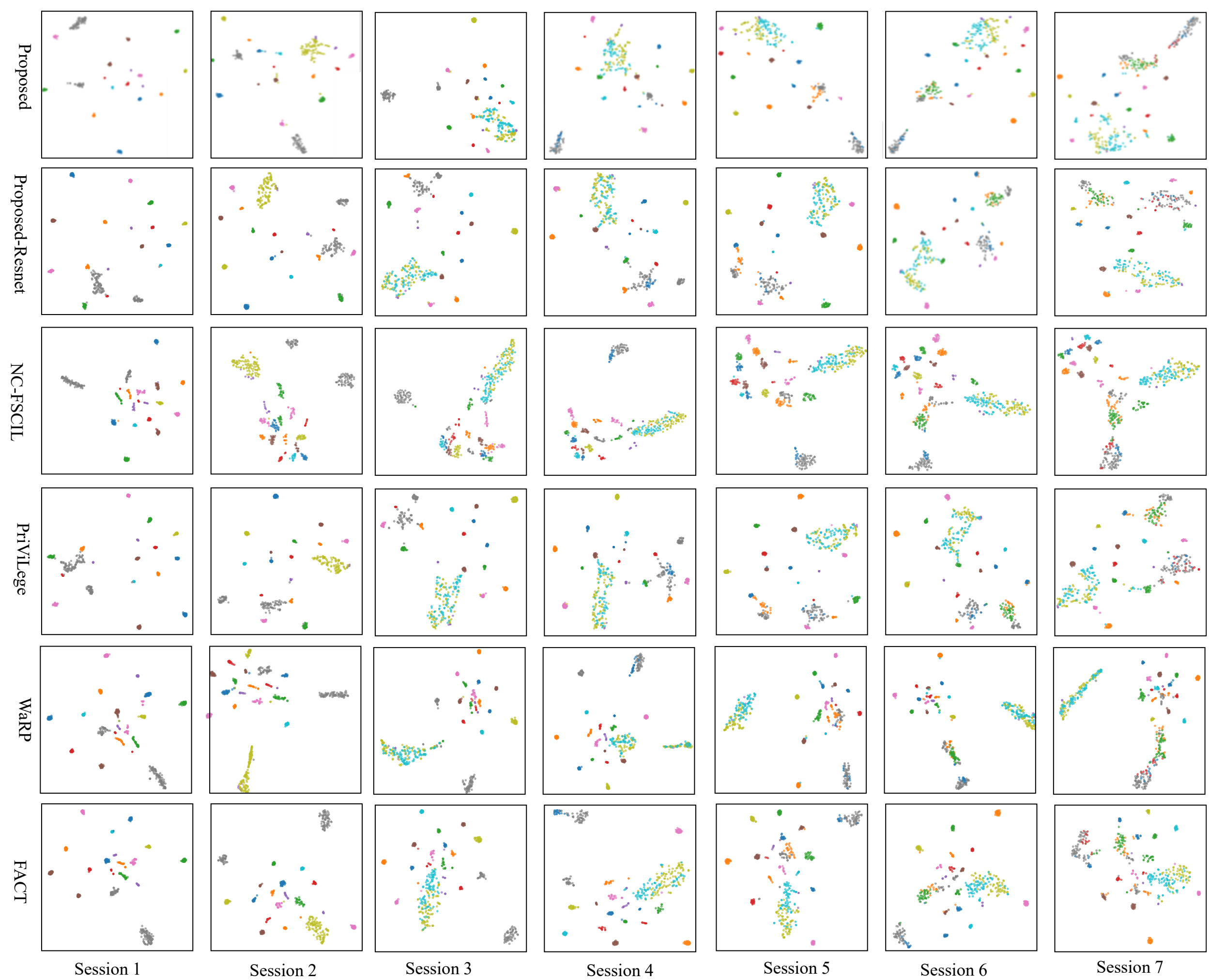

**Fig. 5:** T-SNE visualization of feature distributions across different methods after incremental sessions.

TABLE I
PERFORMANCE COMPARISON OF THE PROPOSED FSCIL METHOD

| Methods | Pub. Year | Accuracy in each session | | | | | | | | |
|---|---|---|---|---|---|---|---|---|---|---|
| | | *Base* ↑ | *Incremental* ↑ | | | | | | | |
| | | $t=0$ | $t=1$ | $t=2$ | $t=3$ | $t=4$ | $t=5$ | $t=6$ | $t=7$ | PD ↓ |
| Proposed | - | 99.69 | 96.15 | 87.45 | 80.64 | 84.58 | 82.27 | 78.14 | **76.46** | **23.23** |
| Proposed-ResNet | - | 99.88 | 93.01 | 95.95 | 86.68 | 80.93 | 78.81 | 77.15 | 71.99 (-4.47) | 27.89 |
| NC-FSCIL | ICLR 23’ | 100.00 | 88.91 | 89.83 | 77.31 | 75.86 | 72.04 | 74.16 | 75.15 (-1.31) | 24.85 |
| PriViLege | CVPR 24’ | 81.06 | 84.47 | 87.37 | 79.42 | 79.73 | 78.38 | 77.15 | 74.80 (-1.66) | 6.26 |
| WaRP | ICLR 23’ | 99.45 | 98.40 | 90.66 | 84.63 | 76.78 | 76.32 | 75.06 | 71.62 (-4.84) | 27.83 |
| FACT | CVPR 22’ | 99.71 | 99.57 | 91.66 | 80.38 | 80.99 | 79.79 | 77.14 | 75.49 (-0.97) | 24.22 |

As shown in Table I, we compare the performance of various baselines in the SAR domain. The variant Proposed-ResNet, which adopts a simpler ResNet-based backbone, exhibits relatively strong performance during early incremental sessions due to its rapid convergence. However, its limited capacity to preserve learned knowledge leads to more severe degradation in later sessions, as reflected in its highest Performance Degradation (PD) of 27.89% among all models. Similarly, NC-FSCIL, which uses a fixed ETF classifier to align features with prototypes, suffers from notable degradation because the gap between SAR features and predefined prototypes makes convergence to the neural-collapse optimum difficult. PriViLege shows a lower PD, but this is partly because its base-session accuracy is already low; therefore, PD alone does not indicate stronger final recognition ability. Its final accuracy remains lower than that of the proposed method, suggesting limited SAR-specific transferability. WaRP and FACT alleviate forgetting through parameter freezing and feature-space reservation, respectively, but they still fall short of the proposed method, which benefits from geometric regularization and neural-collapse-driven cross-domain alignment.

As shown in Fig. 10, each row corresponds to an FSCIL

method, and each column corresponds to an incremental session following the fixed order of newly introduced categories: A10, AT402, Cessna 208, B1B, B52, C130, and F4. The t-SNE visualizations show that the proposed method turns initially ambiguous decision boundaries into better separated clusters, especially in later incremental sessions. This comparison confirms the importance of cross-domain knowledge transfer and geometric regularization for improving class separation and robustness in SAR FSCIL.

### C. Ablation Study

TABLE II
ABLATION EXPERIMENTS OF LOSSES ON THE PROPOSED FSCIL SCHEME

| Loss | | | Accuracy in each session | | | | | | | | | |
|---|---|---|---|---|---|---|---|---|---|---|---|---|
| | | | *Base* ↑ | *Incremental* ↑ | | | | | | | | |
| $\mathcal{L}_{DR}$ | $\mathcal{L}_{orth}$ | $\mathcal{L}_{center}$ | $t=0$ | $t=1$ | $t=2$ | $t=3$ | $t=4$ | $t=5$ | $t=6$ | $t=7$ | PD ↓ | Avg |
| √ | | | 99.69 | 90.86 | **96.70** | 79.65 | 72.46 | 66.91 | 68.63 | 66.81 (-9.65) | 32.88 | 80.21 (-5.46) |
| √ | √ | | 99.69 | 96.15 | 87.45 | 80.54 | 84.42 | 81.60 | 74.25 | 74.44 (-2.02) | 25.25 | 84.82 (-0.86) |
| √ | | √ | 99.69 | 90.66 | 96.82 | 80.01 | 71.19 | 70.55 | 61.38 | 64.38 (-12.08) | 35.31 | 79.34 (-6.34) |
| √ | √ | √ | **99.69** | **96.15** | **87.45** | **80.64** | **84.58** | **82.27** | **78.14** | **76.46** | 23.23 | 85.67 |

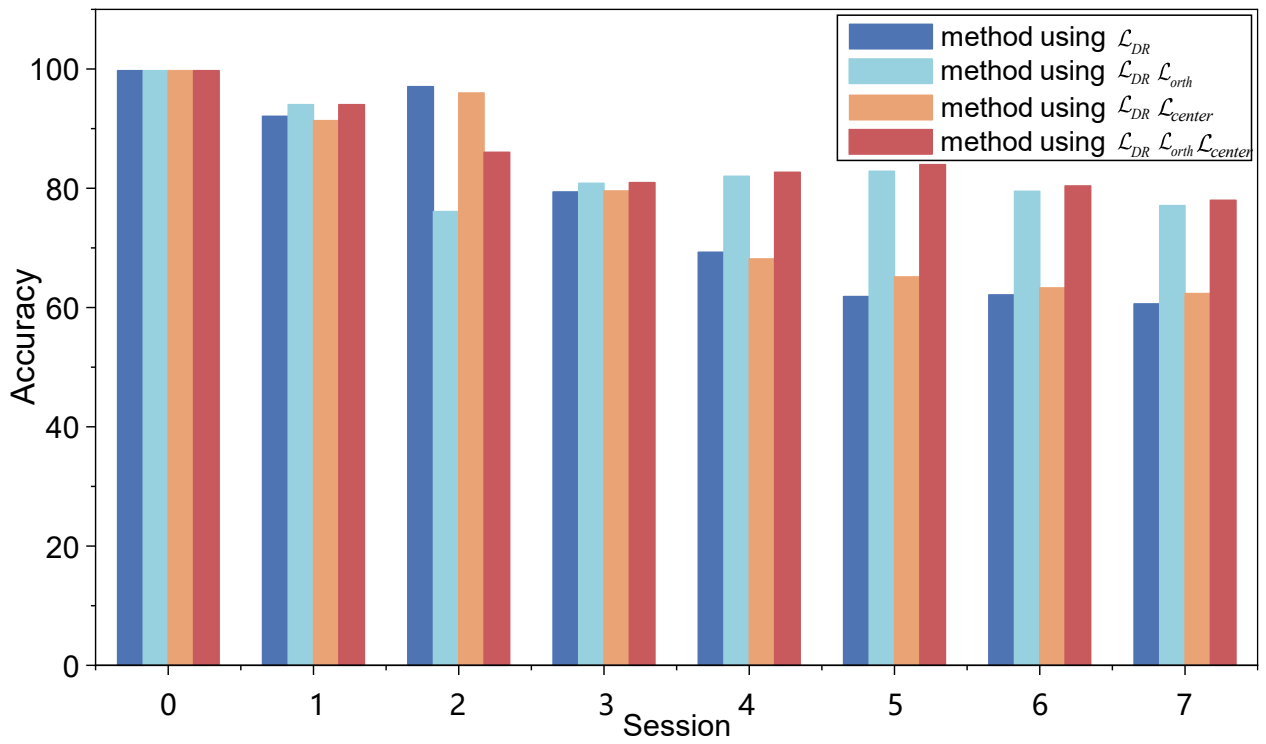

**Fig. 11**: Accuracy comparison in the ablation experiments.

To further validate whether the proposed alignment losses promote neural collapse in SAR features, we conduct a quantitative analysis after the ablation experiments. As detailed in Table II, the ablation study investigates the contribution of each loss component within the proposed SAR FSCIL framework. The variant using only the prototype-alignment loss serves as the baseline: it initially aligns features with class-specific prototypes but fails to preserve inter-class separation when new classes arrive, leading to feature entanglement and forgetting. Adding the PA-based subspace loss improves stability by projecting SAR features toward the refined optical subspace. Adding the subspace-center loss further aligns SAR features with optical subspace centers. Combining these terms produces the most stable performance and supports more reliable convergence toward neural collapse.

Given that both $\mathcal{L}_{orth}$ and $\mathcal{L}_{center}$ are designed to promote neural collapse, we next evaluate whether their inclusion indeed enhances this phenomenon during training. To this end, we adopt three canonical neural collapse metrics— $\Sigma^{\mathcal{NC}_1}$ , $ETF_{std}^{\mathcal{NC}_2}$ , $ETF_{mean}^{\mathcal{NC}_2}$ and $SD^{\mathcal{NC}_3}$ —each corresponding to a distinct sub-phenomenon ( $\mathcal{NC}_1 \sim \mathcal{NC}_3$ ). $\Sigma^{\mathcal{NC}_1}$ measures the collapse of within-class variability ( $\mathcal{NC}_1$ ). $ETF_{std}^{\mathcal{NC}_2}$ and $ETF_{mean}^{\mathcal{NC}_2}$ evaluates whether the class means converge to an ETF structure ( $\mathcal{NC}_2$ ). $SD^{\mathcal{NC}_3}$ quantifies the alignment between classifier weight vectors and class means ( $\mathcal{NC}_3$ ). Together, these metrics provide a comprehensive assessment of how neural collapse evolves during incremental learning and allow us to gauge the effectiveness of our ablation design.

We extend the ablation experiments across the proposed, baseline and two variant methods (Table II), computing neural collapse metrics at each incremental stage. In particular, $\Sigma^{\mathcal{NC}_1}$ evaluates whether the last-layer feature variance within the same class approaches zero, i.e., whether features collapse tightly around their class mean as the within-class covariance vanishes. Formally, $\Sigma^{\mathcal{NC}_1}$ measures the collapse of within-class variance relative to between-class separation, and is computed as:

$$\Sigma^{\mathcal{NC}_1} = \log\left(\frac{\mathrm{Tr}\left(\Sigma_W \Sigma_B^{+}\right)}{C}\right) \tag{31}$$

where $\Sigma_W \in \mathbb{R}^{d \times d}$ is the within-class covariance matrix of the feature representations, $\Sigma_B \in \mathbb{R}^{d \times d}$ is the between-class covariance matrix, $\Sigma_B^{+}$ denotes the Moore–Penrose pseudoinverse of $\Sigma_B$ , $\mathrm{Tr}(\cdot)$ is the trace operator, $C$ is the number of classes, and $d$ is the feature dimension. This metric captures the relative magnitude of within-class variation to between-class variation. A lower $\Sigma^{\mathcal{NC}_1}$ value implies stronger feature collapse within classes, consistent with the theoretical expectation of neural collapse. In the idealized limit, perfect intra-class collapse would yield $\Sigma_W \to 0$ , resulting in $\Sigma^{\mathcal{NC}_1} \to 0$ . Notably, our evaluation focuses exclusively on correctly classified samples, as neural collapse emerges only when misclassification errors vanish. In few-shot incremental learning, misclassified features are randomly scattered and introduce heavy noise, so the metrics are computed solely on correctly classified samples.

We also evaluate the $ETF_{std}^{\mathcal{NC}_2}$ and $ETF_{mean}^{\mathcal{NC}_2}$, which examines whether the class means, after being centered by the global mean, asymptotically align with the vertices of an ETF ( $\mathcal{NC}_2$ ). This condition reflects the emergence of uniformly distributed inter-class geometry. To quantify $\mathcal{NC}_2$, we compute complementary metrics based on the pairwise cosine angles between centered class means and their alignment with the ETF structure.

We first calculate the standard deviation of pairwise cosine angles between all feature means. Let $\mu_c$ and $\mu_{c'}$ be the class means for classes $c$ and $c'$, and let $\mu_G$ be the global mean. We define the cosine between any pairwise feature means as:

$$\cos(\mu_c, \mu_{c'}) = \frac{\langle \mu_c - \mu_G, \mu_{c'} - \mu_G \rangle}{\| \mu_c - \mu_G \| \cdot \| \mu_{c'} - \mu_G \|}, \quad \forall c \neq c' \tag{32}$$

The angular variance ( $ETF_{std}^{\mathcal{NC}_1}$ ) is then computed as:

$$ETF_{std}^{\mathcal{NC}_2} = \text{Std}_{c \neq c'} \left( \cos(\mu_c, \mu_{c'}) \right) \tag{33}$$

A lower $ETF_{std}^{\mathcal{NC}_2}$ indicates a more uniformly distributed angular geometry. We then evaluate the mean pairwise angle among feature means:

$$\bar{\theta}_\mu = \mathbb{E}_{c \neq c'} [\cos(\mu_c, \mu_{c'})] \tag{34}$$

and analogously, the mean pairwise angles between classifier weight vectors $w_c$ and $w_{c'}$ can be calculated as:

$$\bar{\theta}_w = \mathbb{E}_{c \neq c'} [\cos(w_c, w_{c'})] \tag{35}$$

The absolute deviation between the feature and classifier angular means is defined as:

$$ETF_{mean}^{\mathcal{NC}_2} = \left| \bar{\theta}_\mu - \bar{\theta}_w \right| \tag{36}$$

This discrepancy quantifies how closely the learned feature space aligns with the ETF classifier in terms of angular structure. A smaller $ETF_{mean}^{\mathcal{NC}_2}$ implies stronger alignment, indicating that the class-mean geometry of the features better mimics that of the ETF classifier.

We further evaluate the $SD^{\mathcal{NC}_3}$ metric, which assesses the convergence between class means and the corresponding classifier weight vectors in the last layer. This condition captures the self-duality property ( $\mathcal{NC}_3$ ), indicating that feature centers are closely aligned with classifier parameters. To quantify $SD^{\mathcal{NC}_3}$, we measure the Frobenius norm difference between the centered class mean matrix and the classifier weight matrix. Formally, let $\tilde{M} \in \mathbb{R}^{d \times C}$ be the centered class mean matrix and $\tilde{W} \in \mathbb{R}^{d \times C}$ be the centered classifier weights. We compute:

$$SD^{\mathcal{NC}_3} = \left\| \hat{W}^\top - \hat{M}^\top \right\|_F \tag{37}$$

where $\hat{M} = \tilde{M} / \| \tilde{M} \|_F$ and $\hat{W} = \tilde{W} / \| \tilde{W} \|_F$. A smaller value indicates stronger self-duality and closer alignment between the feature space and the classifier space, which is a key characteristic of neural collapse.

TABLE III

NEURAL COLLAPSE METRICS AND ACCURACY IN THE PROPOSED METHOD USING $\mathcal{L}_{DR}$, $\mathcal{L}_{orth}$, $\mathcal{L}_{aux}$

| Session $t$ | $\Sigma^{\mathcal{NC}_1}$ ↓ | $ETF_{std}^{\mathcal{NC}_2}$ ↓ | $ETF_{mean}^{\mathcal{NC}_2} = \left\| \bar{\theta}_\mu - \bar{\theta}_w \right\|$: $\bar{\theta}_\mu$ | $\bar{\theta}_w$ | $ETF_{mean}^{\mathcal{NC}_2}$ ↓ | $SD^{\mathcal{NC}_3}$ ↓ | Acc ↑ |
|---|---|---|---|---|---|---|---|
| 1 | -3.083 | 0.168 | 1.343 | 1.518 | 0.175 | 0.536 | 96.15 |
| 2 | -3.002 | 0.248 | 1.285 | 1.515 | 0.23 | 0.550 | 87.45 |
| 3 | -3.345 | 0.177 | 1.424 | 1.518 | 0.094 | 0.486 | 80.64 |
| 4 | -3.058 | 0.219 | 1.467 | 1.523 | 0.056 | 0.570 | 84.58 |
| 5 | -2.605 | 0.221 | 1.443 | 1.525 | 0.082 | 0.587 | 82.27 |
| 6 | -2.995 | 0.220 | 1.512 | 1.523 | 0.011 | 0.586 | 78.14 |
| 7 | -2.743 | 0.225 | 1.501 | 1.527 | 0.026 | 0.622 | 76.46 |

TABLE IV

NEURAL COLLAPSE METRICS AND ACCURACY IN THE BASELINE METHOD USING $\mathcal{L}_{DR}$

| Session $t$ | $\Sigma^{\mathcal{NC}_1}$ ↓ | $ETF_{std}^{\mathcal{NC}_2}$ ↓ | $ETF_{mean}^{\mathcal{NC}_2} = \left\| \bar{\theta}_\mu - \bar{\theta}_w \right\|$: $\bar{\theta}_\mu$ | $\bar{\theta}_w$ | $ETF_{mean}^{\mathcal{NC}_2}$ ↓ | $SD^{\mathcal{NC}_3}$ ↓ | Acc ↑ |
|---|---|---|---|---|---|---|---|
| 1 | -3.588 (-0.505) | 0.137 (-0.031) | 1.604 | 1.515 | 0.089 (-0.086) | 0.405 (-0.131) | 90.86 (-5.29) |
| 2 | -3.561 (-0.559) | 0.158 (-0.09) | 1.568 | 1.518 | 0.05 (-0.18) | 0.414 (-0.136) | 96.7 (9.25) |
| 3 | -3.673 (-0.328) | 0.121 (-0.056) | 1.591 | 1.52 | 0.071 (-0.023) | 0.399 (-0.087) | 79.65 (-0.99) |
| 4 | -3.034 (0.024) | 0.171 (-0.048) | 1.584 | 1.523 | 0.061 (0.005) | 0.499 (-0.071) | 72.46 (-12.12) |
| 5 | -2.724 (-0.119) | 0.18 (-0.041) | 1.567 | 1.525 | 0.042 (-0.04) | 0.547 (-0.04) | 66.91 (-15.36) |
| 6 | -2.624 (0.371) | 0.186 (-0.034) | 1.564 | 1.523 | 0.041 (0.03) | 0.551 (-0.035) | 68.63 (-9.51) |
| 7 | -2.62 (0.123) | 0.227 (0.002) | 1.575 | 1.529 | 0.046 (0.02) | 0.641 (0.019) | 66.81 (-9.65) |

TABLE V

NEURAL COLLAPSE METRICS AND ACCURACY IN VARIANT METHOD USING $\mathcal{L}_{DR}$, $\mathcal{L}_{orth}$

| Session $t$ | $\Sigma^{\mathcal{NC}_1}$ ↓ | $ETF_{std}^{\mathcal{NC}_2}$ ↓ | $ETF_{mean}^{\mathcal{NC}_2} = \left\| \bar{\theta}_\mu - \bar{\theta}_w \right\|$: $\bar{\theta}_\mu$ | $\bar{\theta}_w$ | $ETF_{mean}^{\mathcal{NC}_2}$ ↓ | $SD^{\mathcal{NC}_3}$ ↓ | Acc ↑ |
|---|---|---|---|---|---|---|---|
| 1 | -3.083 (+0.0) | 0.137 (-0.031) | 1.343 | 1.518 | 0.175 (+0.0) | 0.536 (+0.0) | 96.15 (+0.00) |
| 2 | -3.002 (+0.0) | 0.158 (-0.09) | 1.285 | 1.515 | 0.231 (+0.001) | 0.551 (+0.0) | 87.45 (+0.00) |
| 3 | -3.341 (0.004) | 0.121 (-0.056) | 1.425 | 1.518 | 0.093 (-0.001) | 0.486 (+0.0) | 80.54 (-0.10) |
| 4 | -3.042 (0.016) | 0.171 (-0.048) | 1.469 | 1.523 | 0.055 (-0.001) | 0.572 (+0.002) | 84.42 (-0.16) |
| 5 | -2.609 (-0.004) | 0.171 (-0.05) | 1.444 | 1.526 | 0.082 (+0.000) | 0.588 (+0.001) | 81.60 (-0.67) |
| 6 | -2.909 (0.086) | 0.186 (-0.034) | 1.501 | 1.524 | 0.023 (+0.012) | 0.601 (+0.015) | 74.25 (-3.89) |
| 7 | -3.033 (-0.29) | 0.227 (0.002) | 1.507 | 1.526 | 0.019 (-0.008) | 0.630 (+0.008) | 74.44 (-2.02) |

TABLE VI

NEURAL COLLAPSE METRICS AND ACCURACY IN VARIANT METHOD USING $\mathcal{L}_{DR}$, $\mathcal{L}_{center}$

| Session $t$ | $\Sigma^{\mathcal{NC}_1}$ ↓ | $ETF_{std}^{\mathcal{NC}_2}$ ↓ | $ETF_{mean}^{\mathcal{NC}_2}=\lvert\bar{\theta}_\mu-\bar{\theta}_w\rvert$ | | | $SD^{\mathcal{NC}_3}$ ↓ | Acc ↑ |
|---|---|---|---|---|---|---|---|
| | | | $\bar{\theta}_\mu$ | $\bar{\theta}_w$ | $ETF_{mean}^{\mathcal{NC}_2}$ ↓ | | |
| 1 | -3.564 (-0.481) | 0.138 (-0.030) | 1.605 | 1.515 | 0.090 (-0.085) | 0.406 (-0.130) | 90.66 (-5.49) |
| 2 | -3.549 (-0.547) | 0.157 (-0.091) | 1.570 | 1.518 | 0.051 (-0.179) | 0.414 (-0.136) | 96.82 (9.37) |
| 3 | -3.695 (-0.350) | 0.095 (-0.082) | 1.604 | 1.518 | 0.086 (-0.008) | 0.349 (-0.137) | 80.01 (-0.63) |
| 4 | -3.011 (+0.047) | 0.168 (-0.051) | 1.585 | 1.523 | 0.061 (+0.005) | 0.485 (-0.085) | 71.19 (-13.39) |
| 5 | -2.349 (+0.257) | 0.185 (-0.036) | 1.572 | 1.526 | 0.047 (-0.035) | 0.545 (-0.042) | 70.55 (-11.72) |
| 6 | -2.593 (+0.403) | 0.194 (-0.027) | 1.574 | 1.523 | 0.050 (+0.039) | 0.590 (+0.004) | 61.38 (-16.76) |
| 7 | -2.311 (+0.432) | 0.197 (-0.028) | 1.569 | 1.528 | 0.041 (+0.015) | 0.608 (-0.014) | 64.38 (-12.08) |

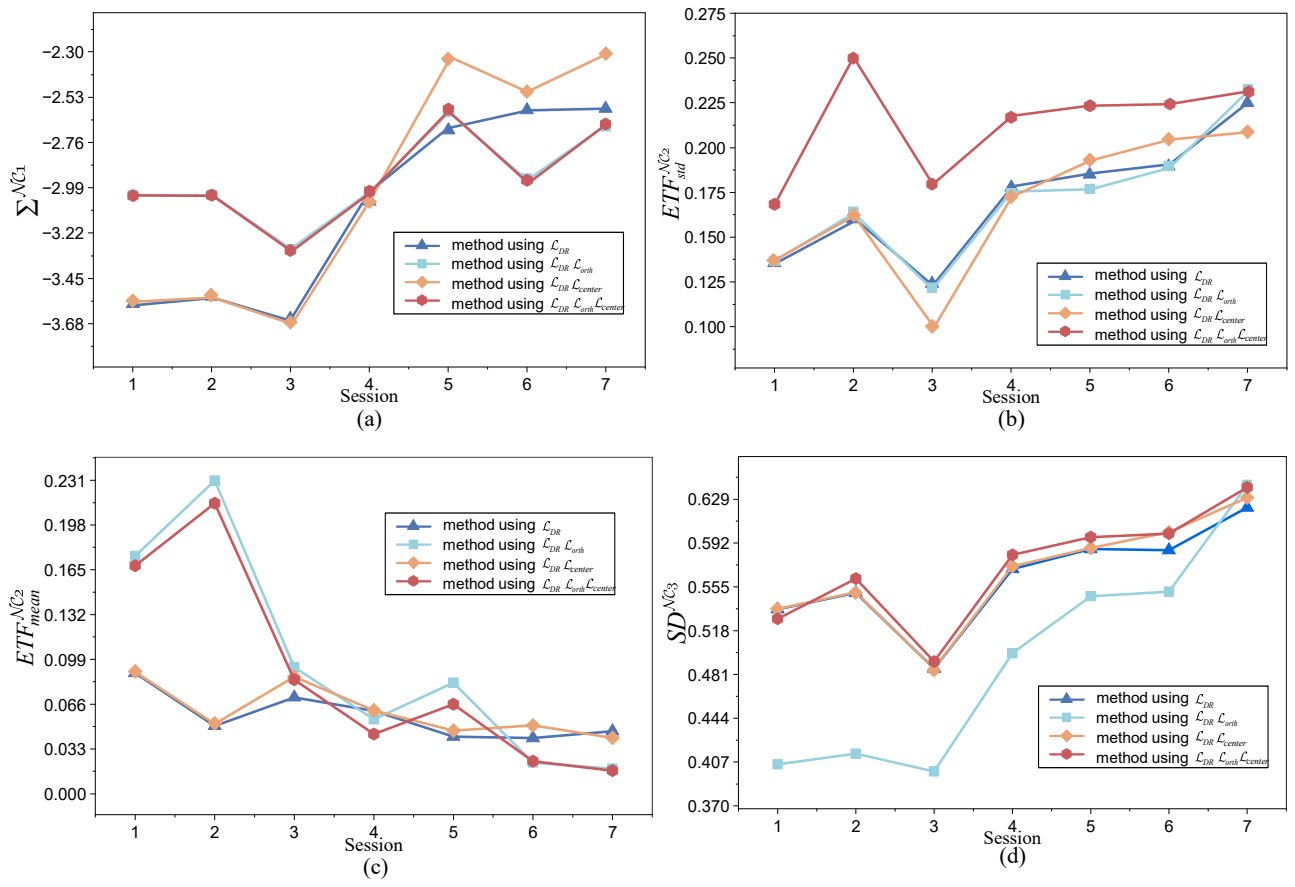

**Fig. 6:** Neural collapse metrics across incremental sessions under different loss configurations. (a) $\Sigma^{\mathcal{NC}_1}$ metrics, (b) $ETF_{std}^{\mathcal{NC}_2}$ metrics, (c) $ETF_{mean}^{\mathcal{NC}_2}$ metrics, and (d) $SD^{\mathcal{NC}_3}$ metrics.

As shown in Fig. 11, the proposed method performs comparably to, or slightly worse than, the baseline in the early incremental sessions (t = 1-3). Therefore, our analysis focuses mainly on the later sessions (t = 4-7), where the trends in both neural-collapse metrics and accuracy are clearer.

Table III reports the neural-collapse metrics and accuracy of the proposed method using the complete loss formulation. For readability, red indicates a value that is more favorable than Proposed for that metric (according to the upward or downward arrow), whereas blue indicates a less favorable value.

Tables IV and V present the results of the baseline method (using only $\mathcal{L}_{DR}$) and the variant method with $\mathcal{L}_{DR}$ and $\mathcal{L}_{orth}$. These comparisons demonstrate that although the baseline maintains initial performance, it suffers from larger neural collapse metrics and severe accuracy degradation in later sessions, while the orthogonal variant alleviates this issue by enforcing tighter class compactness.

Table VI shows the results of the variant method with $\mathcal{L}_{DR}$ and $\mathcal{L}_{center}$. The outcomes indicate that $\mathcal{L}_{center}$ explicitly enhances alignment between SAR features and optical features, leading to improved $ETF_{std}^{\mathcal{NC}_2}$ and more stable accuracy compared with the baseline.

For clarity, Fig. 6 provides a visual comparison of the neural collapse metrics across different methods. Since our loss setting encourages features to lie in orthogonal subspaces, it may increase the angular variability between certain class pairs. In Fig. 6 (b), the $ETF_{std}^{\mathcal{NC}_2}$ of the proposed method is slightly higher than that of the baseline in some sessions, which is an expected side effect of the orthogonal constraints imposed by the $\mathcal{L}_{orth}$ and $\mathcal{L}_{center}$. Also, for methods using $\mathcal{L}_{orth}$, lower $ETF_{mean}^{\mathcal{NC}_2}$ is observed. This metric directly reflects the degree to which the feature means align with a predefined ETF structure. As in Fig. 6 (c), a reduced $ETF_{mean}^{\mathcal{NC}_2}$ implies that the proposed orthogonal constraints push the class means more tightly toward the ETF, enhancing the inter-class separability in angular terms. In other words, even if the $\mathcal{L}_{orth}$ causes some angle variability, it effectively pushes the SAR features into the desired orthogonal partitions, yielding better global collapse of the ETF structure. As in Fig. 6 (d), the $SD^{\mathcal{NC}_3}$ metric further highlights the differences between our approach and the baseline. A lower $SD^{\mathcal{NC}_3}$ indicates that feature means and classifier weights are nearly parallel. Across most incremental sessions, the proposed method shows slightly higher $SD^{\mathcal{NC}_3}$ values, likely due to the orthogonal constraint: by enforcing features to align with fixed orthogonal directions from optical data, the model might initially trade off some direct alignment between features and classifier weights.

Finally, the accuracy of the proposed method in later sessions confirms the practical impact of these geometric improvements. The baseline accuracy drops steeply over time, indicating severe catastrophic forgetting and a misaligned feature space for older classes. In contrast, the proposed method maintains higher accuracy and stronger neural-collapse behavior, suggesting that its decisions are supported by a more robust and geometrically structured feature representation.

## VI. Conclusion

In this paper, we addressed catastrophic forgetting in FSCIL for SAR target recognition. Drawing on recent theoretical advances in neural collapse and ETF-based feature alignment, we proposed a framework that transfers orthogonal geometric structure from data-rich optical imagery to few-shot SAR incremental learning.

Specifically, we first used large-scale optical data to construct orthogonal feature subspaces, where pairwise orthogonal partitions serve as proxy anchors for guiding SAR features toward neural-collapse-like geometry. We then introduced a three-part SAR FSCIL strategy that combines ETF prototype alignment, subspace PA alignment, and subspace-center alignment. This design preserves old-class separability while improving the stability of new-class learning under severe data scarcity.

Overall, the proposed orthogonality-guided alignment framework provides a practical way to transfer structured optical-domain geometry to SAR FSCIL, enabling more robust and interpretable representation learning in low-resource incremental-recognition scenarios.